\documentclass{ieeetj}
\usepackage{cite}
\usepackage{amsmath,amssymb,amsfonts}
\usepackage{graphicx,color}
\usepackage{textcomp}
\usepackage{xcolor}

\def\BibTeX{{\rm B\kern-.05em{\sc i\kern-.025em b}\kern-.08em
    T\kern-.1667em\lower.7ex\hbox{E}\kern-.125emX}}
\AtBeginDocument{\definecolor{tmlcncolor}{cmyk}{0.93,0.59,0.15,0.02}\definecolor{NavyBlue}{RGB}{0,86,125}}

\usepackage{url}
\usepackage{footmisc}
\usepackage{bm}
\newtheorem{rem}{Remark}
\usepackage{subfigure}
\usepackage{ragged2e}
\usepackage{multirow}
\usepackage{utfsym}
\usepackage{amssymb}
\usepackage{makecell}
\usepackage{utfsym}

\usepackage{algpseudocode}
\usepackage{threeparttable}
\usepackage{booktabs}
\usepackage[ruled,vlined,linesnumbered]{algorithm2e}

\def\OJlogo{\vspace{-4pt}$<$Society logo(s) and publication title will appear here.$>$}
\def\seclogo{\vspace{10pt}$<$Society logo(s) and publication title will appear here.$>$}

\def\authorrefmark#1{\ensuremath{^{\textbf{#1}}}}

\usepackage{hyperref}  %hyperref still needs to be put at the end!
\hypersetup{
            colorlinks=true,
            linkcolor=blue,
            anchorcolor=blue,
            citecolor=blue
            }
\begin{document}
\receiveddate{XX Month, XXXX}
\reviseddate{XX Month, XXXX}
\accepteddate{XX Month, XXXX}
\publisheddate{XX Month, XXXX}
\currentdate{XX Month, XXXX}
\doiinfo{XXXX.2022.1234567}

\markboth{}{Author {et al.}}

\title{CTE-MLO: Continuous-time and Efficient Multi-LiDAR Odometry with Localizability-aware Point Cloud Sampling}
\author{Hongming Shen\authorrefmark{1}, Member, IEEE, Zhenyu Wu\authorrefmark{1}, Yulin Hui\authorrefmark{3}, Wei Wang\authorrefmark{1,2}, Graduate Student Member, IEEE, Qiyang Lyu\authorrefmark{1,2}, Tianchen Deng\authorrefmark{2}, Yeqing Zhu\authorrefmark{2}, Bailing Tian\authorrefmark{3}, Member, IEEE, and Danwei Wang \authorrefmark{1,2}, Life Fellow, IEEE}
\affil{Centre for Advanced Robotics Technology Innovation (CARTIN), Nanyang Technological University, Singapore, 639798 Singapore}
\affil{School of Electrical and Electronic Engineering, Nanyang Technological University, Singapore, 639798 Singapore}
\affil{School of Electrical and Information Engineering, Tianjin University, Tianjin, 300072 China}
\corresp{Corresponding author: Zhenyu Wu (email: ZHENYU002@e.ntu.edu.sg).}
\authornote{This research is supported by the National Research Foundation (NRF),
Singapore, under the NRF Medium Sized Centre scheme (CARTIN), the
Agency for Science, Technology and Research (A*STAR) under its National
Robotics Programme with Grant No. M22NBK0109, and National Research
Foundation, Singapore and Maritime and Port Authority of Singapore under its Maritime Transformation Programme (Project No. SMI-2022-MTP-04).
\\Any opinions, findings and conclusions or recommendations expressed in
this material are those of the author(s) and do not reflect the views of
National Research Foundation, Singapore and Maritime and Port Authority
of Singapore.}

\begin{abstract}
In recent years, LiDAR-based localization and mapping methods have achieved significant progress thanks to their reliable and real-time localization capability.
Considering single LiDAR odometry often faces hardware failures and degeneracy in practical scenarios, Multi-LiDAR Odometry (MLO), as an emerging technology, is studied to enhance the performance of LiDAR-based localization and mapping systems.
However, MLO can suffer from high computational complexity introduced by dense point clouds that are fused from multiple LiDARs, and the continuous-time measurement characteristic is constantly neglected by existing LiDAR odometry.
This motivates us to develop a Continuous-Time and Efficient MLO, namely CTE-MLO, which can achieve accurate and real-time estimation using multi-LiDAR measurements through a continuous-time perspective.
In this paper, the Gaussian process estimation is naturally combined with the Kalman filter, which enables each LiDAR point in a point stream to query the corresponding continuous-time trajectory using its time instants.
A decentralized multi-LiDAR synchronization scheme is also devised to combine points from separate LiDARs into a single point cloud without the primary LiDAR assignment.
Moreover, with the aim of improving the real-time performance of MLO without sacrificing robustness, a point cloud sampling strategy is designed with the consideration of localizability.
To this end, CTE-MLO integrates synchronization, localizability-aware sampling, continuous-time estimation, and voxel map management within a Kalman filter framework, which can achieve high accuracy and robust continuous-time estimation within only a few linear iterations.
The effectiveness of the proposed method is demonstrated through various scenarios, including public datasets and real-world applications.
Exhaustive benchmarks on public dataset show that CTE-MLO is demonstratively competitive compared to other State-of-the-art (SOTA) methods in accuracy, robustness, and efficiency.
Furthermore, CTE-MLO has undergone comprehensive validation across various scenarios (such as port, campus, and forest) and on diverse platforms (a full-size truck, an autonomous sweeper, and an agile Micro Aerial Vehicle (MAV)), covering a total test area exceeding $3.6km^2$, which affirm the applicability of CTE-MLO in industrial and field scenarios. A demonstration video is available at \url{https://youtu.be/Q29PGPitHUI}, and the code is also available at \url{https://github.com/shenhm516/CTE-MLO} to benefit the community.
\end{abstract}



\maketitle
\section{Introduction}
\IEEEPARstart{L}{ocalization} is a fundamental task in developing autonomous mobile robots. LiDAR odometry has become one of the mainstream localization solutions in Global Navigation Satellite System (GNSS)-denied environments, which has the potential to enable extensive industrial applications, such as construction\cite{doi:10.1126/scirobotics.abp9758, 10616135}, autonomous driving\cite{10579830,eventdetection}, and underground mine exploration\cite{10576057,subtCERBERUS, subt-data61}.
During the last decade, LiDAR odometry has achieved remarkable localization and mapping accuracy thanks to the direct depth measurement of LiDAR, which is immune to scene illumination and texture changes.
However, many existing approaches\cite{LOAM,DLO,FLOAM} treat continuous LiDAR point streaming as an instantaneous measurement through the motion undistortion technique.
The discrete-time estimation of LiDAR-only odometry suffers from the low frame rate (e.g., 10Hz) of the LiDAR, making it challenging to deploy on agile platforms that require high-frequency state estimation for control and planning feedback.
To overcome this issue, many existing approaches fuse LiDAR measurements with the Inertial Measurement Unit (IMU) to provide high-frequency state estimation in a loosely coupled\cite{LOCUS,LOCUS2,DLIO} or tightly coupled\cite{FAST-LIO2,LIO-SAM,9826792} manner.
Unfortunately, recent research\cite{Point-LIO,Traj-LO} indicates that LiDAR-inertial odometry can not provide stable state estimation in certain extreme cases, such as vibration and aggressive motions.
This motivates us to develop a continuous-time LiDAR odometry that treats the LiDAR point cloud as a continuous-time point stream and utilizes each LiDAR point in the stream to constrain a continuous-time trajectory corresponding to its time domain.

With further consideration of the degeneracy issues\cite{degeneracy-zhangji,xicp,MGLT,MM4MM}, extra onboard sensors (such as visual camera\cite{Zhang2017,zhang2018laser,LVI-SAM,R3LIVE,LU2022107521}, GNSS\cite{GLIO,LIO-SAM,9762805}, encoder\cite{9561254,10342258,xicp}, and Ultra-wide Bandwidth (UWB) anchors\cite{VIRAL-Fusion,10504602,LIRO}) are generally adopted to provide constraints to the degeneracy direction of LiDAR odometry through the multi-sensor fusion technique.
However, visual cameras can degenerate significantly or even become unobservable in conditions of poor illumination and low texture, and GNSS cannot work in an indoor or dense urban area due to the multi-path effect\cite{9562037}. Besides, encoders can only be deployed on legged or wheel robots, and UWB anchors rely on installation and calibration.
To improve the robustness of state estimation for field robots, a fraction of works \cite{MLOAM,MLOAM-SS} focus on robust state estimation using multi-LiDAR measurements in defense of LiDAR odometry.
With the continuous reduction in the cost, weight, and power consumption of LiDAR in recent years, multi-LiDAR localization systems can now be deployed not only on large-size robotic platforms such as autonomous driving vehicles \cite{10579830,M-LIO} and excavators\cite{doi:10.1126/scirobotics.abp9758, 10616135} but also on small-size robotic platforms (e.g., unmanned ground vehicles\cite{MLOAM,MLOAM-SS}, legged robots\cite{doi:10.1126/scirobotics.adi9641}, even MAV\cite{zhao2021super,VIRAL-Fusion}) in such real-world industrial and field applications.
However, the number of points provided by multiple LiDARs worsens the real-time performance of MLO, which significantly increases the dimension of the estimation problem.
The estimation of LiDAR odometry is a solve-associate-solve loop that implies the MLO can benefit more from the correct association than the nonlinear iterations.
As an extreme example, the ego-motion between two LiDAR point clouds with exact point association can be estimated in a closed-form\cite{ICP} without iteration.
Based on this idea, the closed-form solution of the Extended Kalman Filter (EKF) is introduced as a prior of association in each iteration to replace the monotonic cost function descent in nonlinear optimization (e.g., Gaussian Newton, Levenberg-Marquardt), thereby improving the real-time performance of MLO.
With further considering that the number of points provided by multiple LiDARs is a trade-off between robustness and real-time performance, we are motivated to develop a point cloud sampling strategy with the consideration of localizability.
Hence, in this paper, an EKF-based Continuous-Time and Efficient MLO, so-called CTE-MLO, is developed with a localizability-aware point cloud sampling strategy.

The main contributions of this paper are listed as follows:
\begin{itemize}
    \item Considering the inherent continuous-time characteristics of the LiDAR point stream, we extend the potential of LiDAR-only odometry and develop a continuous-time MLO that enables real-time and precise trajectory estimation using pure LiDAR measurements without any additional sensor fusion.
    \item A Gaussian process representation is adopted to describe the continuous-time trajectory, which enables a natural combination of the continuous-time LiDAR odometry with the Kalman filter. To the best of our knowledge, CTE-MLO is the first filter-based continuous-time MLO, which can solve the continuous-time trajectory estimation problem efficiently within a few linear iterations.
    \item A decentralized multi-LiDAR synchronization scheme is designed to combine multiple LiDAR measurements using the LiDAR splitting technique, which is robust to arbitrary LiDAR failure. Considering the large number of points provided by synchronized multi-LiDAR observation, which preserves a high level of geometric constraint redundancy.
    An efficient localizability-aware point cloud sampling method is proposed to improve the real-time performance of MLO, which quantifies the localizability contribution for each point, and ensures localizability using a minimal number of points with high localizability contribution.
    \item A full-fledged continuous-time MLO framework is proposed, which integrates decentralized synchronization, localizability-aware sampling, continuous-time estimation, and voxel map management. Exhaustive benchmark comparison on public datasets shows that CTE-MLO achieves competitive accuracy and efficiency when compared with SOTA MLO methods.
    Real-world industrial and field applications on various platforms, including a full-scale truck, an autonomous sweeper, and an agile autonomous exploration MAV, are deployed to demonstrate the performance and industrial applicability of CTE-MLO.
\end{itemize}

\section{Related Works}
To develop a continuous-time and efficient MLO, related SOTA methods are reviewed with the subject of LiDAR odometry, continuous-time LiDAR odometry, and MLO.
\subsection{LiDAR Odometry:} LiDAR odometry is designed to estimate the robot’s motion between LiDAR acquisitions using the point cloud registration technique represented by Iterative Closest Point (ICP) \cite{ICP}, Generalized Iterative Closest Point (GICP) \cite{GICP}, and Normal Distribution Transformation (NDT) \cite{NDT}.
One of the most popular LiDAR SLAM algorithms LOAM\cite{LOAM} provides a standard framework for LiDAR odometry, which achieves real-time ego-motion estimation through the scan-to-scan matching and low-drift registration through the scan-to-map matching.
The odometry and mapping framework proposed in LOAM has been widely adopted in recent research.
Several techniques used in LOAM, such as feature extraction, scan-to-scan matching, and motion estimation, are improved by current SOTA LiDAR SLAM methods.
In DLO\cite{DLO}, a direct LiDAR odometry is developed to provide real-time pose estimation using the minimally preprocessed dense point cloud without feature extraction.
A pruned GICP registration algorithm is designed in\cite{DLO} to accelerate the scan-to-scan and scan-to-map process with an adaptive keyframe strategy.
In FLOAM\cite{FLOAM}, the scan-to-scan matching in LOAM\cite{LOAM} is replaced by the assumption of a uniform motion model.
LeGO-LOAM\cite{LeGO-LOAM} improves the accuracy and real-time performance of LOAM by segmenting the point from the ground.
To further improve the localization and mapping performance, MULLS\cite{MULLS} is proposed by segmenting the points in the scan with more detailed types (ground, facade, roof, pillar, beam, and vertex) when compared with LeGO-LOAM\cite{LeGO-LOAM}.
The aforementioned methods achieve LiDAR-based estimation using the GICP registration or variants (e.g. performing registration using the principal direction of the covariance in GICP).
In recent years, a few methods attempting to return to the roots, which introduce the classic ICP registration to develop a LiDAR odometry in a concise formulation.
Open3d-SLAM \cite{jelavic2022open3d} provides a fully LiDAR SLAM solution, including the ICP-based estimation, loop closure, and map management, leveraging a highly performant point cloud library Open3D \cite{Open3D}.
However, the performance of ICP registration highly relies on the accuracy of the initial guess, because it imposes point association in the nearest distance.
KISS-ICP\cite{KISS-ICP} solves this problem by combining the classical ICP registration with an adaptive matching strategy, which yields a lightweight LiDAR-only localization system in defense of point-to-point ICP.
Besides LiDAR odometry based on ICP, GICP, and their variants, NDT provides a more lightweight estimation and map management for LiDAR odometry \cite{hdl-localization,NDT-LOAM}.
Different from GICP registration, which performs the normal distribution approximation using an exact nearest neighbor search, NDT approximates normal distribution through lightweight voxelization clustering.
However, NDT-based LiDAR odometry \cite{hdl-localization,NDT-LOAM} tends to cluster voxels in a pre-defined (fixed) size, which reduces the adaptability to variational scenarios.
To overcome this, in BALM\cite{balm}, a novel adaptive voxelization mapping structure is developed using the hash data structure, and a theoretical framework for performing LiDAR Bundle Adjustment (BA) is also proposed.
Although BA has achieved good performance in the visual odometry field\cite{Zhang2017,VINS-MONO,orbslam3,S-MSCKF,Deng_2024_CVPR}, for LiDAR odometry which has 10–100 times of points than visual odometry, registration-based LiDAR odometry is still the mainstream, while LiDAR BA is commonly applied in the fields of non-real-time estimation and mapping, such as fine pose graph optimization\cite{BALM2} and consistent map building\cite{HBA}.
For tasks with high-frequency estimation requirements, e.g. feedback control for agile platform, current mainstream LiDAR odometry achieves real-time estimation through fusion with IMU which neglects the high-bandwidth characteristic of the LiDAR point streaming.

\subsection{Continuous-time LiDAR Odometry}
The aforementioned LiDAR odometry treats LiDAR scans as instantaneous measurements by employing motion undistortion techniques and utilizes the undistorted point cloud for the registration process.
However, the motion undistortion relies on the scan-to-scan matching results or uniform motion assumption, which introduces the additional measurement noise into the undistorted point cloud.
To solve this problem, a series of approaches investigate continuous-time LiDAR odometry with the idea of taking into account the motion during scanning by defining a continuous-time trajectory.
In MARS\cite{MARS}, a continuous-time LiDAR odometry is proposed to estimate the B-spline trajectory and remove the motion distortion of the LiDAR scan with a non-rigid registration.
Thanks to the continuous-time B-spline representation, in \cite{CLINS,SLICT2,CLIC}, a serial continuous-time trajectory estimation system is developed by combining asynchronous sensor measurements (e.g., camera, IMU) with the continuous-time LiDAR scan.
However, according to the definition of B-spline\cite{B-spline}, the formulation of the B-spline trajectory is determined by a set of control points, which includes control points not in the current time domain.
Estimation of future control points leads to redundant computations for B-spline-based continuous-time SLAM systems.
To solve this issue, linear interpolation is employed to model the continuous movement straightforwardly.
CT-ICP\cite{CT-ICP} and KISS-ICP\cite{KISS-ICP} show great trajectory estimation results by incorporating the motion undistortion into the registration using linear interpolation.
The uniform motion assumption implicated in linear interpolation leads CT-ICP\cite{CT-ICP} and KISS-ICP\cite{KISS-ICP} not robust enough to sudden changes of orientation or fast acceleration motion.
In Traj-LO\cite{Traj-LO}, a piecewise linear interpolation is adopted to minimize the gap between uniform motion assumption and aggressive motion.
In\cite{GPGN}, the Gaussian Process Gauss-Newton (GPGN) optimization method is derived to enable continuous-time estimation, while \cite{steam} extends the GPGN to the $\text{SE}(3)$ space.
Currently, there only exists limited research focus on the Gaussian process continuous-time LiDAR-only odometry.
In \cite{Picking_up_Speed}, a continuous-time LiDAR odometry is designed to estimate the Gaussian process trajectory for a novel LiDAR type, called Frequency-Modulated Continuous-Wave (FMCW) LiDAR which can measure the relative velocity of each measured point via the Doppler effect.
Unfortunately, the application of FMCW LiDAR is still relatively limited in the industry platforms.
\subsection{Multi-LiDAR Odometry}
Considering that single LiDAR odometry frequently suffers from hardware failure and degeneracy, MLO has become an emerging research topic in SLAM.
MLOAM\cite{MLOAM} is a representative MLO, which has been proven to have better performance compared to single-LiDAR odometry methods.
In MILIOM\cite{MILIOM}, a tightly coupled multi-LiDAR-inertial odometry is proposed by combining multiple LiDAR measurements through a general synchronization scheme.
In SLICT\cite{SLICT} and SLICT2\cite{SLICT2}, the authors extend MILIOM\cite{MILIOM} to continuous-time multi-LiDAR-inertial odometry with linear interpolation and B-spline, respectively.
However, the synchronization strategy adopted in \cite{MLOAM,MILIOM,SLICT,SLICT2} is centralized, which can not provide stable estimation in the situation of primary LiDAR failure.
An intuitive solution to this problem is to develop a decentralized MLO.
In \cite{MLOAM-SS}, a decentralized MLO framework is developed to realize robust state estimation exploits information from multiple Field of View (FOV)-limited LiDARs.
However, decentralized MLO has to estimate the state or extrinsic for each LiDAR, which significantly increases the computational complexity of the state estimation process.
On the mapping side, the dense point cloud acquired from multiple LiDARs poses a challenge to the efficiency of the map management process.
In most conventional MLO \cite{MLOAM, MILIOM, MLOAM-SS}, KD-Tree is introduced as the mapping structure to accelerate the K-NN querying process.
However, conventional KD-Tree does not support incremental updates, which means a compulsory rebuild process is required to maintain the global map.
To solve this problem, incremental KD-Tree (iKD-Tree) \cite{cai2021ikd} and incremental voxel (iVox) \cite{FASTER-LIO} are developed to accelerate the mapping process, which maintains the global map incrementally.
Although iKD-Tree and iVox achieve efficient data insertion, the time complexity of the K-NN search is still related to the size of the KD-Tree and the number of points in the voxel, respectively.
In \cite{voxelmap}, an efficient adaptive voxel mapping algorithm is proposed by combining the octree with the hash data structure, which achieves efficient data association by clustering points in each voxel into normal distributions.
To future improve the mapping efficiency, SLICT\cite{SLICT} and SLICT2\cite{SLICT2} perform incremental updates on each voxel in the voxel map, enabling incremental updates and efficient query of the global map for MLO.
In view of the aforementioned analysis, only a few MLO methods \cite{SLICT,SLICT2} considered the continuous measurement characteristics of LiDAR, which all heavily rely on centralized multi-LiDAR synchronization and IMU measurements.
For pure MLO\cite{MLOAM,MLOAM-SS}, the real-time performance is always worsened by the dense point cloud acquired from multiple LiDARs, the high-dimension state to be estimated, or inefficient map management.
To overcome these problems, in this paper, CTE-MLO is developed to achieve a continuous-time and efficient MLO in a tightly coupled manner by integrating decentralized synchronization, continuous-time estimation, localizability-aware point cloud sampling, and efficient voxel map management.

\section{Gaussian Process Continuous-Time LiDAR Odometry} \label{Sec: Continuous-Time LiDAR Odometry}
A continuous-time LiDAR-only odometry is in charge of estimating a trajectory $\mathbf{x}(t), t \in \left[t_k, t_{k+1}\right)$ with the observation from the LiDAR point cloud, where an arbitrary state vector $\mathbf{x} \in \mathbf{x}(t)$ is assumed to follow a Gaussian distribution.
A Gaussian process is intuitively adopted to describe the continuous-time trajectory $\mathbf{x}(t)$ with a mean function ${\mathbf{\bar x}}(t)$ and a covariance function ${\bm{\sigma }}_p(t,t')$.
\begin{equation}
{\mathbf{x}}(t) \sim {\mathcal{GP}}\left( {{\mathbf{\bar x}}(t),{\bm{\sigma}}_p(t,t')} \right) \label{Eq: GP of x}
\end{equation}
where the covariance function involves two time variables, $t$ and $t'$, to account for cross-temporal relations.

The continuous-time LiDAR-based localization and mapping problem can be formulated as maximizing the posterior probability of the trajectory $\mathbf{x}(t)$ with LiDAR measurements $\mathbf{z}_{t_k}$ plus an initial state $\mathbf{\check{x}}_{t_k}$.
\begin{equation}
{\bf{\hat x}}(t) = \mathop {\arg \max }\limits_{{\bf{x}}(t)} p\left( {{\bf{x}}(t)|{\mathbf{z}_{t_k}},{{\bf{\check{x}}}_{t_k}}} \right) \label{Eq: MAP}
\end{equation}
where $p(\cdot)$ denotes the probability function. 
For a measurement obtained at time $t_i\in\left[t_k,t_{k+1}\right)$, the measurement model is defined to be
\begin{equation}
{{\mathbf{z}}_{t_i}} = h\left({{\bf{x}}(t_i)} \right) + {\mathbf{w}_{t_i}},{\mathbf{w}_{t_i}} \sim {\mathcal{N}}({\mathbf{0}},{{\bm{\sigma }}_{R,t_i}})
\end{equation}
where $h(\cdot)$ denotes the observation function, and ${\bm{\sigma}}_{R,t_i}$ is the covariance of the measurement noise $\mathbf{w}_{t_i}$.

With the Gaussian approximation, the Maximum a Posteriori (MAP) problem defined in (\ref{Eq: MAP}) can be simplified as a least square optimization problem.
\begin{equation}
\mathop {\min }\limits_{{\bf{x}}(t)} \underbrace {\left\| {{\bf{\bar x}}({t_k}) - {{\bf{\check{x}}}_{{t_k}}}} \right\|_{{\bm\sigma}_{P,{t_k}}^{ - 1}}^2 + \sum\limits_{i = 1}^M {\left\| {{{\bf{z}}_{{t_i}}} - h\left( {{\bf{\bar x}}({t_i})} \right)} \right\|_{{\bm\sigma}_{R,{t_i}}^{ - 1}}^2} }_V \label{Eq: lso}
\end{equation}
where $M$ is the number of LiDAR measurements, $\mathbf{\bar x}(t_k)$ is the mean function corresponding to the trajectory $\mathbf{x}(t)$ at time $t=t_k$, and $\mathbf{x}(t_k)$ is subject to a Gaussian distribution $\mathbf{x}(t_k) \sim \mathcal{N}(\mathbf{\bar x}(t_k), {\bm \sigma}_{P,{t_k}})$.

Inspired by \cite{Anderson2015}, the Gaussian Process is assumed to be generated by Linear, Time-varying (LTV) Stochastic Differential Equations (SDE).
\begin{equation}
{\bf{\dot x}}(t) = {{\mathbf{F}}}(t){\mathbf{x}}(t) + {{\mathbf{G}}}(t){\mathbf{w}}(t) \label{Eq: LTVSDE}
\end{equation}
where $\mathbf{F}(t)$ and $\mathbf{G}(t)$ are time-varying system matrices, $\mathbf{w}(t)\sim\mathcal{GP}(\mathbf{0},{\bm\sigma}_{Q}\delta(t-t'))$ is a stationary, zero-mean, white-noise Gaussian process with power-spectral density matrix ${\bm\sigma}_{Q}$, and $\delta(\cdot)$ denotes the Dirac delta function. 

Under the LTV SDE assumption, the mean function of the Gaussian process $\mathbf{x}(t)$ can be expressed by adopting a basis function representation \cite{maybeck1982stochastic}:
\begin{equation}
\begin{aligned}
&\mathbf{\bar x}(t) = {\bm{\Phi}}(t,t_k)\mathbf{\bar x}(t_k)\\
&\begin{aligned}
{{\bm{\sigma }}_P}(t,t') &= {\bf{\Phi }}(t,{t_k}){{\bm{\sigma }}_{P}(t_k,t_k)}{{\bf{\Phi }}^ \top }(t',{t_k}) \\&+ \int_{{t_k}}^{\min(t,t')} {{\bf{\Phi }}(t,s)} {\bf{G}}(s){{\bm\sigma}_{Q}}{{\bf{G}}^ \top }(s){{\bm{\Phi }}^ \top }(t',s)ds
\end{aligned}
\end{aligned}\label{Eq: transition function}
\end{equation}
where ${\bm{\Phi}}(t,t_k)$ is the transition matrix of the LTV system.
Substitute (\ref{Eq: transition function}) into (\ref{Eq: lso}), the cost function $V$ of the localization and mapping problem (\ref{Eq: lso}) can be rewritten as:
\begin{equation}
\begin{aligned}
&\begin{aligned}
V &=
 \left\| {{\bf{\bar x}}({t_k}) - {{\bf{\check x}}_{{t_k}}}} \right\|_{{\bm{\sigma }}_{P,{t_k}}^{ - 1}}^2 + \sum\limits_{i = 1}^M {\left\| {{{\bf{z}}_{{t_i}}} - h\left( {{\bf{\Phi }}({t_i},{t_k}){\bf{\bar x}}({t_k})} \right)} \right\|_{{\bm{\sigma }}_{R,{t_i}}^{ - 1}}^2} \\
&= \left\| {{\bf{\bar x}}({t_k}) - {{\bf{\check x}}_{{t_k}}}} \right\|_{{\bm{\sigma }}_{P,{t_k}}^{ - 1}}^2 + \left\| {{\bf{z}} - {\bf{h}}} \right\|_{{\bm{\sigma }}_R^{ - 1}}^2
\end{aligned}\\
&{\bf{z}} = \left[\cdots, {{\bf{z}}_{{t_i}}^\top}, \cdots\right]^\top,{\bf{h}} = \left[\cdots,{{h^\top}\left( {{\bf{\Phi }}({t_i},{t_k}){\bf{\bar x}}({t_k})} \right)},\cdots\right]^\top\\
&{{\bm{\sigma }}_R} = \text{Diag}\left( {\left[ { \cdots ,{{\bm{\sigma }}_{R,{t_i}}}, \cdots } \right]} \right)
\label{Eq: cost function}
\end{aligned}
\end{equation}
where $\text{Diag}(\cdot)$ denotes the diagonal operation.

To minimize the cost function defined in (\ref{Eq: cost function}) with an initial guess $\mathbf{\check x}_{t_k}$, a first-order approximate is adopted to update $\mathbf{\check x}_{t_k}$ by a small perturbation $\mathbf{\tilde x}_{t_k}$.
% \begin{equation}
\begin{align}
&\mathbf{\bar x}(t_k) = \mathbf{\check x}_{t_k} + \mathbf{\tilde x}_{t_k} \label{Eq: state update}\\
&\begin{aligned}
&{{{\bf{\tilde x}}}_{{t_k}}} = \mathop {\arg \min }\limits_{{{{\bf{\tilde x}}}_{{t_k}}}} {\left\| {{{{\bf{\tilde x}}}_{{t_k}}}} \right\|_{{\bm{\sigma}}_{P,{t_k}}^{ - 1}}^2 + \left\| {{{\bf{z}}} - {{\bf{h}}_{\mathbf{\check x}}} - {\left.{\bf{H}}\right|_{\mathbf{\check x}_{t_k}}}{\bm\Phi}{{{\bf{\tilde x}}}_{{t_k}}}} \right\|_{{\bm{\sigma }}_R^{ - 1}}^2} \label{Eq: SQP}\\
& {\bf{h}}_{\mathbf{\check{x}}} = \left[\cdots,{{h^\top}\left( {{\bf{\Phi }}({t_i},{t_k}){\mathbf{\check x}}({t_k})} \right)},\cdots\right]^\top\\
&{\bm\Phi} = \left[\cdots, {\bm\Phi}^\top(t_i,t_k), \cdots\right]^\top
\end{aligned}
\end{align}
% \end{equation}
where $\mathbf{H}$ is the Jacobian matrix of $\bf h$ with respect to $\left[\cdots, \mathbf{\bar x}^\top(t_i),\cdots\right]^\top$.
The optimization problem defined in (\ref{Eq: SQP}) is a Standard Quadratic Programming (SQP) problem, which has a closed-form solution by taking the derivative with respect to $\mathbf{\tilde x}_{t_k}$.
\begin{equation}
\underbrace {\left( {{{\bf{\Phi }}^ \top }{{\bf{H}}^ \top }{\bm{\sigma }}_R^{ - 1}{\bf{H\Phi }} + {\bm{\sigma}}_{P,t_k}^{ - 1}} \right)}_{\text{Hessian Matrix}}{\bf{\tilde x}}_{t_k} = {{\bf{\Phi }}^ \top }{\bf{H}}^ \top {\bm{\sigma }}_R^{ - 1}\left( {{\bf{z}} - {\bf{h}}} \right) \label{Eq: linear GN}
\end{equation}

Since the derivation follows the Gauss-Newton approach \cite{GPGN}, the covariance can be expressed as the inverse of the Hessian matrix in (\ref{Eq: linear GN}).
\begin{equation}
{{\bm{\sigma }}_P}({t_k},t_k) = {\left( {{{\bf{\Phi }}^ \top }{{\bf{H}}^ \top }{\bm{\sigma }}_R^{ - 1}{\bf{H\Phi }} + {\bm{\sigma }}_{P,t_k}^{ - 1}} \right)^{ - 1}} \label{Eq: updated cov}
\end{equation}
The Gaussian process $\mathbf{x}(t)\sim\mathcal{GP}(\mathbf{\bar x}(t), {\bm \sigma}(t_k,t))$ can be estimated by substitute $\mathbf{\bar x}(t_k)$ (updated in (\ref{Eq: state update}) using the closed-form solution $\mathbf{\tilde x}(t_k)$ of SQP (\ref{Eq: linear GN})) and the updated covariance ${{\bm{\sigma }}_P}({t_k},t_k)$ (\ref{Eq: updated cov}) into (\ref{Eq: transition function}).

\subsection{Combining Gaussian Process with Kalman Filter} \label{Sec: Combining Gaussian Process with Kalman Filter}
With the derivation of (\ref{Eq: GP of x})-(\ref{Eq: updated cov}), the key to estimating the Gaussian process $\mathbf{x}(t)$ is to solve the perturbation $\mathbf{\tilde x}_{t_k}$ and the covariance matrix ${\bm \sigma}_P(t_k,t_k)$ with the initial guess $\mathbf{\check x}_{t_k}$ and ${\bm \sigma}_{P,t_k}$, which process can be realized using the Kalman filter.
To facilitate the derivation, an intermediate matrix $\mathbf{H}'=\mathbf{H}\bm{\Phi}$ is defined, which is the Jacobian matrix of $\mathbf{h}$ with respect to $\mathbf{\bar x}(t_k)$.
The Gauss-Newton equations (\ref{Eq: linear GN}) can be rewritten as:
\begin{equation}
{\bf{\tilde x}}_{t_k} = \underbrace {\underbrace {{{\left( {{\bf{H}}{'^ \top }{\bm{\sigma }}_R^{ - 1}{\bf{H}}' + {\bm{\sigma }}_{P,{t_k}}^{ - 1}} \right)}^{ - 1}}}_{{{\bm{\sigma }}_P}({t_k},{t_k})}{\bf{H}}{'^ \top }{\bm{\sigma }}_R^{ - 1}}_{{{\bf{K}}_{t_k}}}\left( {{\bf{z}} - {\bf{h}}} \right) \label{Eq: new GN}
\end{equation}
where $\mathbf{K}_{t_k}$ is equivalent to the Kalman gain matrix with the Sherman-Morrison-Woodbury identity\cite{Sherman1950AdjustmentOA,woodbury1950inverting}.

Hence, Gaussian process estimation using the Kalman filter is summarized as follows:
\begin{enumerate}
    \item Get the initial guess $\mathbf{\check x}_{t_k}$ and ${\bm \sigma}_{P,t_k}$ through (\ref{Eq: transition function}) which is equivalent to the state prediction process of the Kalman filter.
    \item Update the covariance matrix ${\bm\sigma}_P(t_k,t_k)$ and get the Kalman gain matrix $\mathbf{K}_{t_k}$:
    \begin{equation}
        \begin{aligned}
        &{\bm\sigma}_P(t_k,t_k)={{{\left( {{\bf{H}}{'^ \top }{\bm{\sigma }}_R^{ - 1}{\bf{H}}' + {\bm{\sigma }}_{P,{t_k}}^{ - 1}} \right)}^{ - 1}}} %\label{Eq: kalman cov matrix}
        \\&\mathbf{K}_{t_k} = {\bm\sigma}_P(t_k,t_k)\mathbf{H}'^\top{\bm\sigma}_R^{-1} \label{Eq: kalman gain matrix}
        \end{aligned}
    \end{equation}
    \item Update the mean vector $\mathbf{\bar x}(t_k)$ by substituting (\ref{Eq: new GN}) and (\ref{Eq: kalman gain matrix}) into (\ref{Eq: state update}): 
    \begin{equation}
        \mathbf{\bar x}(t_k) = \mathbf{\check x}_{t_k} + \mathbf{K}_{t_k}(\mathbf{z}-\mathbf{h}) \label{Eq: state update ekf}
    \end{equation}
    \item Step 2-3 iterates until convergence or exceeding the maximum number of iterations. At each iteration, the initial guess is updated according to $\mathbf{\check{x}}_{t_k} = \mathbf{\check{x}}_{t_k} + \mathbf{\tilde{x}}_{t_k}$, with the covariance given by ${\bm{\sigma}}_{P,{t_k}} = {\bm{\sigma }}_{P}({t_k},{t_k})$.
    \item Estimate the Gaussian process $\mathbf{x}(t)$ by substituting $\mathbf{\bar x}(t_k)$ and ${\bm\sigma}_P(t_k,t_k)$ into (\ref{Eq: transition function}).
\end{enumerate}

\subsection{Specialize in Continuous-time LiDAR Odometry}
In this section, a continuous-time LiDAR-only odometry is developed through the state prediction and state update process of the Kalman filter, which are described in more detail below. Consider a continuous-time LiDAR odometry, the state of the Gaussian process is defined as:
\begin{equation}
    {{\bf{x}}(t)} = {\left[ {{\bf{t}}^\top(t) ,{\bm{\phi}}^ \top(t) ,{\bf{v}}^\top(t),  {\bm{\omega}}^\top(t), {\bf{a}}^\top(t)} \right]^ \top } \label{Eq: state vector}
\end{equation}
where $\mathbf{t}(t)$ is the trajectory of translation, $\bm{\phi}(t)$ is the axis-angle representation of the rotation matrix trajectory ${\bf R}(t) \in \text{SO(3)}$, ${\bf v}(t)$ is the velocity trajectory, ${\bm \omega}(t)$ is the angular speed trajectory, and ${\bf a}(t)$ is the linear acceleration trajectory.

\subsubsection{State Prediction}
With the definition of the Gaussian process $\mathbf{x}(t)$ in (\ref{Eq: state vector}), the LTV SDE (\ref{Eq: LTVSDE}) can be specified as:
\begin{equation}
\begin{aligned}
&{\mathbf{\dot t}}(t) = {\bf{v}}(t),{\mathbf{\dot v}}(t) = {\mathbf{R}(t)\mathbf{a}(t)}, \bm{\dot\phi}(t)  = {\bm{\omega}}(t)\\
&{\bm{\dot \omega }}(t) = {{\bf{w}}_\omega }(t),{\bf{\dot a}}(t) = {{\bf{w}}_a}(t)\label{Eq: kinematic model}
\end{aligned}
\end{equation}
where $\mathbf{w}_a(t)$ and $\mathbf{w}_\omega(t)$ are the white noise Gaussian processes of acceleration and angular speed.
For a LiDAR-only odometry, the kinematic model (\ref{Eq: kinematic model}) is adopted instead of the IMU prediction formulation in the state prediction process, and the initial guess $\mathbf{\check{x}}_{t_k}, {\bm \sigma}_{P,t_k}$ can be predicted by introducing (\ref{Eq: kinematic model}) into the general solution of LTV SDE (\ref{Eq: transition function}).
\subsubsection{State Update} \label{sec: state update}
The state update process is in charge of solving the Gauss-Newton equation (\ref{Eq: new GN}) using the continuous-time LiDAR measurements.
For point $\mathbf{p}_i$ ($t_i \in \left[t_{k},t_{k+1}\right)$), the objective of the LiDAR matching process is to construct the state-dependent observation function through geometric constraints between the point $\mathbf{p}_i$ and the voxel map $\mathbb{M}$ (details can be found in Section \ref{Sec: Voxel Map Management}). 
The point-to-voxel matching problem can be formulated as a Maximum Likelihood Estimation (MLE) problem.
\begin{equation}
\max L{(^W}{{\bf{p}}_i}|{\mathbb{M}}) \buildrel \Delta \over = \min  - \log (L{(^W}{{\bf{p}}_i}|{\mathbb{M}}))\label{Eq: MLE}
\end{equation}
where $L{(^W}{{\bf{p}}_i}|{\mathbb{M}})$ represents the likelihood of observing a point $^W{\mathbf{p}_i}$ belonging to the voxel map ${\mathbb{M}}$, and $^W{\mathbf{p}_i}$ is the transformed point defined as
\begin{equation}
^W{{\bf{p}}_i} = {\bf{\bar R}}({t_i}){{\bf{p}}_i} + {\bf{\bar t}}({t_i}) \label{Eq: transformed point}
\end{equation}
where $\{ {\bf{\bar R}}({t_i}),{\bf{\bar t}}({t_i})\} \in {\bf{\bar x}}(t_i)$, and ${\bf{\bar x}}(t_i)= {\bf{\Phi }}(t_i,{t_k}){{{\bf{\bar x}}}({{t_k}})}$.

Following our previous work\cite{PGO-LIOM,TIE-Multi-SLAM}, the likelihood $L{(^W}{{\bf{p}}_i}|{\mathbb{M}})$ can be simplified through the Gaussian Mixture Model (GMM).
\begin{equation}
\begin{aligned}
&L{(^W}{{\bf{p}}_i}|{\mathbb{M}}) = L{(^W}{{\bf{p}}_i}|{\mathbf{m}_j})=\frac{1}{{{{\left[ {{{\left( {2\pi } \right)}^3}\left| {{\bm{\sigma }}_j^m} \right|} \right]}^{\frac{1}{2}}}}}\exp \left( {-D} \right)\\
&- \log \left( {L{(^W}{{\bf{p}}_i}|{{\bf{m}}_j})} \right) = -\underbrace{\log \left( {\frac{1}{{{{\left[ {{{\left( {2\pi } \right)}^3}\left| {{\bm{\sigma }}_j^m} \right|} \right]}^{\frac{1}{2}}}}}} \right)}_{\textbf{independent of } \mathbf{x}(t)} + D
\end{aligned} \label{Eq: GMM}
\end{equation}
where $D=\frac{1}{2}\left( {{}^W{{\bf{p}}_i} - {\bm{\mu }}_j^m} \right)^\top{\bm{\sigma }}_j^{m, - 1}{\left( {{}^W{{\bf{p}}_i} - {\bm{\mu }}_j^m} \right)}$, ${\bm\mu}_j^m$ is the mean vector of the voxel $\mathbf{m}_j \sim \mathcal{N}({\bm\mu}_j^m, {\bm \sigma}_j^m)$, and $\mathbf{m}_j$ is the voxel corresponding to ${^W{\bf{p}}_i}$. 

According to (\ref{Eq: GMM}), the MLE problem (\ref{Eq: MLE}) can be simplified as a least-square optimization problem:
\begin{equation}
\min {\left( {{}^W{{\bf{p}}_i} - {\bm{\mu }}_j^m} \right)^\top}{{\bm \sigma}_j^m}^{-1}\left( {{}^W{{\bf{p}}_i} - {\bm{\mu }}_j^m} \right) \label{Eq: p2v gmm}
\end{equation}
Adopting local geometry approximated by GMM will affect the stability of MLE if the covariance matrix ${{\bm \sigma}_j^m}$ does not have an inverse. A stabilization technique is adopted in
this work, which simplifies (\ref{Eq: p2v gmm}) through the principal component analysis.

\begin{equation}
% \begin{aligned}
\min {\left( {{}^W{{\bf{p}}_i} - {\bm{\mu }}_j^m} \right)^\top}{\bf n}_j^m{\bf n}_j^{m,\top}\left( {{}^W{{\bf{p}}_i} - {\bm{\mu }}_j^m} \right) \label{Eq: p2v pca}
\end{equation}
where ${\bf n}_j^m$ is the normal vector of the voxel $\mathbf{m}_j \sim \mathcal{N}({\bm\mu}_j^m, {\bm \sigma}_j^m)$.

Hence, the observation model of the $i$-th continuous-time LiDAR matching factor can be defined as
\begin{equation}
{z_{t_i}} = 0,{h\left({\bm \Phi}(t_i,t_k)\mathbf{\bar x}(t_k)\right)} = {\bf{n}}_j^{m,\top}({}^W{{\bf{p}}_i} - {\bm{\mu }}_j^m) \label{Eq: observation model}
\end{equation}

From the definition of the observation model (\ref{Eq: observation model}), the Jacobian of ${h\left({\bm \Phi}(t_i,t_k)\mathbf{\bar x}(t_k)\right)}$ over the mean function of the Gaussian process $\mathbf{\bar x}({t_k})$ can be calculated as follows:
\begin{equation}
% \small{
\begin{aligned}
&{{\bf{H}}'_i} = \left.\frac{{h\left({\bm \Phi}(t_i,t_k)\mathbf{\bar x}(t_k)\right)}}{{\partial {{\bf{\bar x}}(t_k)}}}\right|_{\mathbf{\check x}_{t_k}}
=\left[ {\bf{H}'}_i^{{{\bf{t}}}},{\bf{H}'}_i^{{{\bf{R}}}},{\bf{H}'}_i^{{{\bf{v}}}},{\bf{H}'}_i^{{{\bf{\omega }}}},{\bf{H}'}_i^{{{\bf{a}}}} \right]\\
&{\bf{H}'}_i^{{{\bf{t}}}} = {\bf{n}}_j^{m, \top },{\bf{H}'}_i^{{{\bf{v}}}} = {\bf{n}}_j^{m, \top }\Delta t,{\bf{H}'}_i^{{{\bf{a}}}} = \frac{1}{2}{\bm{\eta}}_j\Delta {t^2}\\
&{\bf{H}'}_i^{{{\bf{R}}}} =  - {\bm{\eta}}_j\left( {{{\left\lfloor {\exp \left( {{{\left\lfloor {{{\bm{\check \omega}}_{t_k}}\Delta t} \right\rfloor }_ \times }} \right){{\bf{p}}_i}} \right\rfloor }_ \times } + \frac{1}{2}{{\left\lfloor {{{\bf{\check a}}_{t_k}}} \right\rfloor }_ \times }\Delta {t^2}} \right)\\
&{\bf{H}'}_i^{{{\bm{\omega}}}} =  - {\bm{\eta}}_j\exp \left( {{{\left\lfloor {{{\bm{\check \omega }}_{t_k}}\Delta t} \right\rfloor }_ \times }} \right){\left\lfloor {{{\bf{p}}_i}} \right\rfloor _ \times }\Delta t
% &{\bm{\eta}}_j = {\bf{n}}_j^{m, \top }{{\bf{R}}_{t_k}}
\end{aligned} \label{Eq: Jaco}
% }
\end{equation}
where ${\bm{\eta}}_j = {\bf{n}}_j^{m, \top }{{\bf{\check R}}_{t_k}}$, ${{\left\lfloor \cdot \right\rfloor }_ \times }$ returns the skew-symmetric matrix of a vector, $\Delta t_i = t_i - t_{k}$, and $\mathbf{H}'_i$ is the $i$-th row of $\mathbf{H}'$. The right Lie derivative model \cite{6727494} is introduced to derivate the Jacobian in (\ref{Eq: Jaco}).

Substitute the continuous-time LiDAR observation model (\ref{Eq: observation model}) and the analytic Jacobians (\ref{Eq: Jaco}) into (\ref{Eq: kalman gain matrix}) and (\ref{Eq: state update ekf}), the mean vector and covariance matrix of the state vector $\mathbf{x}(t_k) \in \mathbf{x}(t)$ can be estimated in the state update process of Kalman filter, and the Gaussian process $\mathbf{x}(t)$ can also be estimated by substitute $\mathbf{x}(t_k)\sim\mathcal{N}(\mathbf{\bar x}(t_k),{\bm{\sigma}}(t_k,t_k))$ into (\ref{Eq: transition function}).

\begin{rem} \label{rem: continue time}
\textit{(Continuous-time LiDAR odometry) With the derivation in Section \ref{Sec: Combining Gaussian Process with Kalman Filter}, the Gaussian process estimation problem (\ref{Eq: lso}) can be solved by combining the continuous-time LTV SDE solution (\ref{Eq: transition function}) with the Kalman filter.
Conventional LiDAR odometry\cite{LOAM,DLO,FLOAM,LeGO-LOAM,MULLS,MLOAM-SS,MLOAM} generally performs state estimation with Kalman filter or optimization in a discrete formulation which treats LiDAR point cloud as an instantaneous measurement through the deskew technique.
However, different from instantaneous sensor suites, such as camera and IMU, LiDAR is naturally a continuous-time sampling sensor that captures LiDAR points continuously and accumulates LiDAR points as a point cloud $\mathbb{P}$ before publishing.
In this paper, a continuous-time LiDAR odometry is developed by estimating the Gaussian process represented continuous-time trajectory in a Kalman filter formulation, which can handle the distortion of the LiDAR scan during the estimation process.}
\end{rem}

\section{Point Cloud Synchronization and Sampling}
This section gives a detailed description of the multi-LiDAR point cloud preprocessing, which is in charge of providing observations for the continuous-time LiDAR odometry.
\subsection{Decentralized Multi-LiDAR Synchronization} \label{Sec: Decentralized Multi-LiDAR Synchronization}
\begin{figure}[!t]\centering
\includegraphics[width=\linewidth]{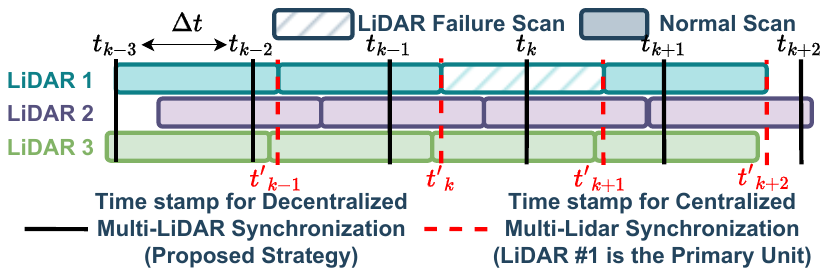}
\caption{\justifying Decentralized multi-LiDAR synchronization scheme. For centralized synchronization, no point can merge in a period ${\left[t'_{k},t'_{k+1}\right)}$ due to the primary LiDAR failure (LiDAR $\#1$). The decentralized synchronization scheme can obtain synchronized point clouds even when only one LiDAR is functioning.}\label{Fig: Synchronization}
% \vspace{-2em}
\end{figure}

\begin{figure*}[!t]
    \centering	
    \subfigure[Point stream from different LiDAR.]{
        \includegraphics[width=0.48\linewidth]{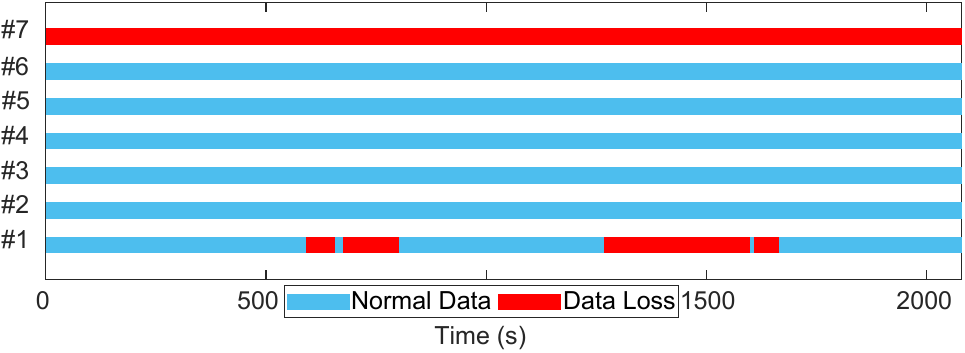}
    }
    % \quad
    \subfigure[{Performance of centralized synchronization\cite{SLICT2} when using different LiDAR as the primary LiDAR.}]{
        \includegraphics[width=0.48\linewidth]{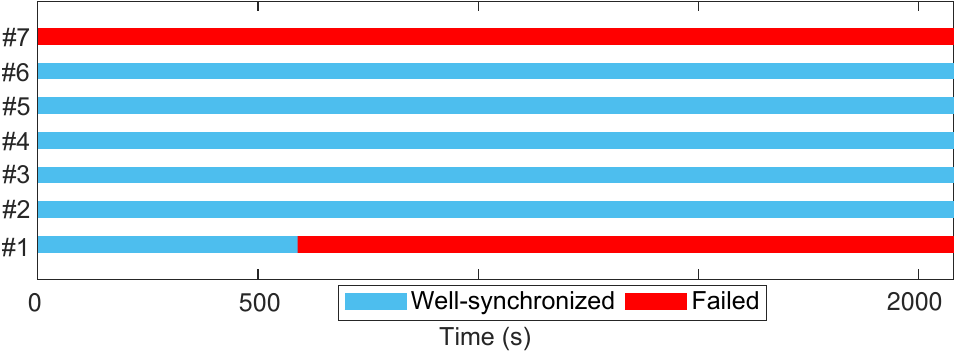} 
        \label{Fig: centralized sync}
    }
    \subfigure[Synchronized point stream. The two figures in the left column illustrate the effectiveness of the proposed decentralized synchronization strategy under LiDAR data loss conditions. The fusion of LiDAR point clouds from different LiDAR (shown in different colors), that pertain to the same period, can provide denser and almost panorama horizontal FOV in a slight time interval $\Delta t=0.01s$. The two figures in the right column showcase the point stream downsampled using the proposed localizability-aware point cloud sampling.]{
        \includegraphics[width=\linewidth]{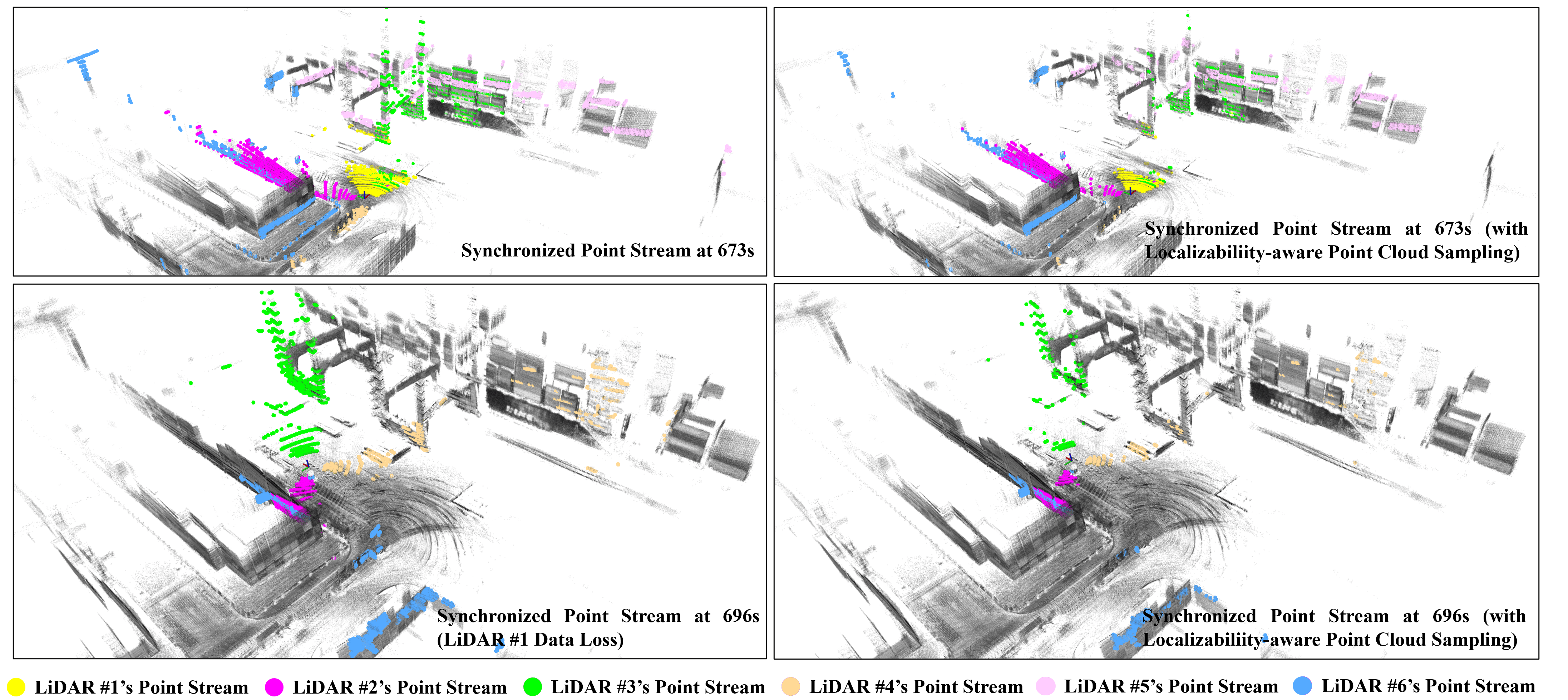} 
        \label{Fig: port-CloudMerge}
    }
    % \vspace{-1em}
    \caption{\justifying Demonstration of proposed multi-LiDAR synchronization and sampling strategy under sequence \textit{Port 01}. LiDAR $\#1$ experienced data loss at $591s-654s, 675s-801s, 1269-1597s$, and $1609-1664s$ due to overly high temperatures, and the data of LiDAR $\#7$ is totally lost due to hardware failure.} \label{Fig: Data loss}
    % \vspace{-1em}
\end{figure*}

To perform joint estimation across multiple LiDARs in a tightly coupled manner, data streams have to be synchronized.
In practice, all LiDARs are connected to the onboard computer through high-speed Ethernet cables. A popular robotics communication middleware, ROS \cite{ros1}, is adopted to synchronize all LiDAR topics into the Coordinated Universal Time (UTC) with the LiDAR drivers.
Hence, in this section, synchronization is considered in the data stream aspect.
For most existing multi-LiDAR synchronization schemes\cite{MLOAM,MILIOM,SLICT,SLICT2}, multi-LiDAR measurements are synchronized to the end time of the primary LiDAR (so-called centralized synchronization), which is not robust to the primary LiDAR data loss.
As shown in Fig. \ref{Fig: Synchronization}, when the primary LiDAR fails (in period $[t'_{k},t'_{k+1})$), the centralized synchronization scheme is unable to synchronize other LiDARs' points to the primary LiDAR.
Considering the unstable LiDAR data stream in practical applications, commonly caused by sensor overheating or unreliable wiring, a decentralized multi-LiDAR synchronization strategy is proposed, which treats LiDAR messages as a continuous-time point stream.
For each point in the LiDAR point cloud message, we can simultaneously obtain its coordinate $\mathbf{p}_i$ (in LiDAR frame) and time stamp $t_i$ (UTC) through the LiDAR driver.
At time $t_{k}$, all points obtained from each LiDAR in a period $\left[t_{k},t_{k}+\Delta t \right)$ are merged into a point cloud $\mathbb{P}$.
The proposed method achieves multi-LiDAR synchronization with a fixed time interval $\Delta t$ using the LiDAR scan splitting technique, which implies the proposed synchronization strategy is totally decentralized and robust to arbitrary LiDAR failure.
To illustrate the effectiveness, as shown in Fig. \ref{Fig: Data loss}, the proposed multi-LiDAR synchronization strategy is evaluated in a real-world autonomous driving scenario (sequence \textit{Port 01} in Table \ref{Tab: Data sequences information} is adopted for testing).
The conventional centralized synchronization strategy\cite{SLICT2}, shown in Fig. \ref{Fig: centralized sync}, can lead to point cloud synchronization failure and result in localization failure when the pre-defined primary LiDAR failure (LiDAR $\#1$ experienced data loss at $591-654s, 675-801s, 1269-1597s$, and $1609-1664s$ due to overly high temperatures, and the data of LiDAR $\#7$ is totally lost due to the wiring problem).
The proposed synchronization strategy is completely decentralized, which splits the point stream from each LiDAR with a fixed time interval $\Delta t = 0.01s$ and merges them into a point cloud $\mathbb{P}$.
As illustrated in Fig. \ref{Fig: port-CloudMerge}, the fusion of multiple LiDAR point streams provides denser observations compared to a single split LiDAR point stream (represented by different colors in Fig. \ref{Fig: port-CloudMerge}). The proposed decentralized synchronization strategy combines multiple point streams, each with a limited FOV, into a single dense LiDAR point cloud with an almost panoramic horizontal FOV at a frequency of 100 Hz.
\subsection{Localizability-aware Point Cloud Sampling} \label{sec: Localizability-aware Point Cloud Sampling}
Multi-LiDAR SLAM methods generally suffer from high latency due to a high level of point redundancy which leads to expensive data association and estimation.
The point cloud sampling strategy is a trade-off between the real-time performance and the stability of LiDAR odometry. 
Removing too many points from the point cloud may lead to accuracy reduction or even degeneration (the point-to-voxel registration problem (\ref{Eq: MLE}) degenerated into a non-convex optimization problem).
Hence, in this paper, a localizability-aware point cloud sampling method is proposed to remove point redundancy with the evaluation of LiDAR degeneracy.
It is worth noting that the conventional point sampling methods \cite{9561262,xicp} require knowledge about point-to-map correspondence relationships.
Getting point-to-map correspondence for MLO is expensive as the point-to-map matching has to be recomputed in every iteration.
This motivates us to develop a point cloud sampling method without prior knowledge of the point-to-map matching information.
\begin{figure}[!t]
	\centering	
	\subfigure[{Translation Perspective}]{
		\includegraphics[width=0.9\linewidth]{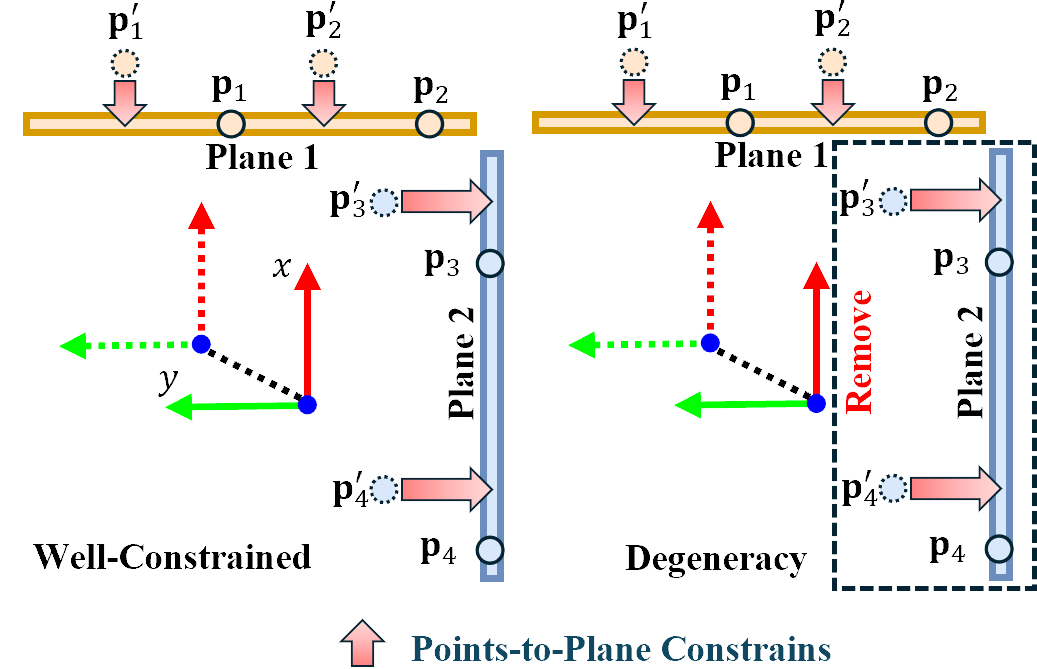}
		\label{Fig: Translation Localizability}		
	}
	% \quad
	\subfigure[{Rotation Perspective}]{
		\includegraphics[width=0.9\linewidth]{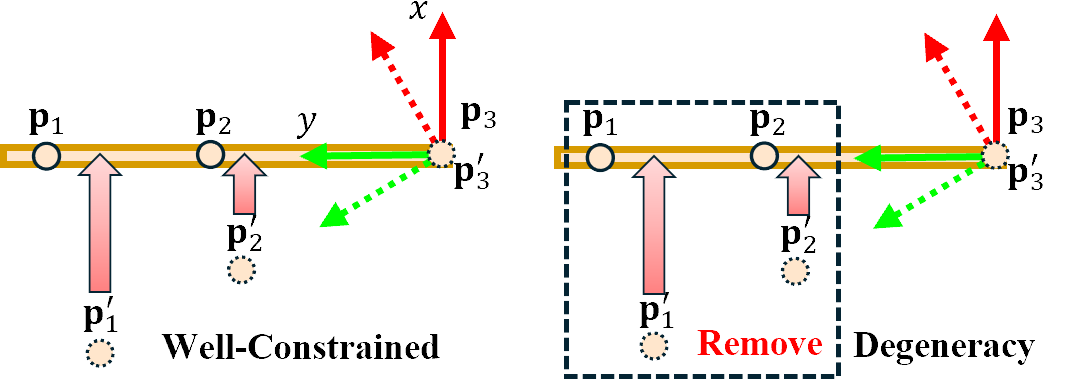} 
		\label{Fig: Rotaion Localizability}
	}
	\caption{{An illustration of the localizability concept in 2D.}} 
    \label{Fig: Localizability}
    % \vspace{-1em}
\end{figure}
\begin{rem} \label{rem: Definition of Localizability}
    \textit{(Definition of Localizability) In this paper, localizability is a metric to determine if the geometry observations can well constrain the LiDAR-based estimation.
    To illustrate the definition of localizability, a 2D example is provided in Fig. \ref{Fig: Localizability}. Fig. \ref{Fig: Translation Localizability} illustrates an environment consisting of two perpendicular planes that well constrain the LiDAR odometry. 
    For example, if Plane 2 in Fig. \ref{Fig: Translation Localizability} is removed, the LiDAR odometry will degenerate in the normal direction of Plane 2, which means $\mathbf{p}_3$ and $\mathbf{p}_4$ contribute to the localizability in the normal direction of their corresponding plane (Plane 2).
    For the rotation perspective, as shown in Fig. \ref{Fig: Rotaion Localizability}, $\mathbf{p}_1$ and $\mathbf{p}_2$ contribute localizability, and the rotation estimation will be dominated by observation noise (so-called rotation degeneracy) when $\mathbf{p}_1$ and $\mathbf{p}_2$ are removed.}
\end{rem}
To facilitate the LiDAR degeneracy analysis, we substitute the LiDAR observation model (\ref{Eq: observation model}) into (\ref{Eq: lso}) and neglect the predictive constraints (irrelevant to LiDAR observations).
\begin{equation}
\mathop {\min }\limits_{{\bf{x}}(t)} {\left\|\mathbf{h}\right\|^2_{{\bm\sigma}_{R}^{-1}} } \buildrel \Delta \over =  \mathop {\min }\limits_{{\bf{x}}(t)} {\sum\limits_{i = 1}^{M} {\left\| {{\bf{n}}_j^{m,\top}({\bf{\bar R}}(t_i){{\bf{p}}_i} + {\bf{\bar t}}(t_i) - {\bm{\mu }}_j^m)} \right\|} ^2_{\bm \sigma_{R,t_i}^{-1}}} \label{Eq: p2v registration}
\end{equation}
Referring to the derivations in Section \ref{Sec: Continuous-Time LiDAR Odometry}, the Gaussian process Kalman filter solves (\ref{Eq: p2v registration}) by iteratively updating $\bf \bar x$ with the error state $\bf \tilde x$.
\begin{equation}
\begin{aligned}
&\mathop {\min }\limits_{{{{\bf{\tilde x}}}}} \left\| \mathbf{h}_{\mathbf{\check x}} +\mathbf{H}'\mathbf{\tilde x} \right\|_{{\bm\sigma} _R^{ - 1}}^2 \buildrel \Delta \over = \\
&\mathop {\min }\limits_{{{{\bf{\tilde x}}}}} {{{\bf{\tilde x}}}^\top}{{\bf{H}}'^\top}{\bm\sigma}_R^{ - 1}{\bf{H'\tilde x}} + 2{\bf h}_{\mathbf{\check x}}^\top{\bm\sigma} _R^{ - 1}{\bf{H'\tilde x}} + {\bf h}_{\mathbf{\check x}}^\top{\bm\sigma} _R^{ - 1}{\mathbf{h}_{\mathbf{\check x}}}
\end{aligned}
% \end{array}
\label{Eq: SQP2}
\end{equation}
The optimal solution of (\ref{Eq: SQP2}) can thus be obtained by solving $
{\raise0.7ex\hbox{${\partial \left\| {{\bf h}_{\mathbf{\check x}} + {\bf{H'\tilde x}}} \right\|_{\bm\sigma _R^{ - 1}}^2}$} \!\mathord{\left/
 {\vphantom {{\partial \left\| {h({\bf{\hat x}}) + {\bf{H\tilde x}}} \right\|_{\sigma _R^{ - 1}}^2} {\partial {\bf{\tilde x}}}}}\right.\kern-\nulldelimiterspace}
\!\lower0.7ex\hbox{${\partial {\bf{\tilde x}}}$}} = 0$.
\begin{equation}
\underbrace {{{\bf{H}'}^ \top }{\bm\sigma} _R^{ - 1}{\bf{H}'}}_{\bf{A}}{\bf{\tilde x}} =  - {{\bf{H}'}^ \top }{\bm\sigma} _R^{ - 1} \mathbf{h}_{\mathbf{\check x}} \label{Eq: Hession}
\end{equation}
where $\bf A$ corresponding to the Hessian matrix of (\ref{Eq: SQP2}). 
Intuitively, the inverse of ${\bm\sigma}_R$ provides weight for each LiDAR observation.
As the trick used in \cite{FAST-LIO2}, which set all LiDAR observations to have the same weight, the Hessian matrix can be simplified as:
\begin{equation}
{\bf{A}}' = {{\bf{H}'}^ \top }{\bf{H}'}=
\left[{\begin{array}{*{10}{c}}
     {{\bf A}'_{tt}}   &{{\bf A}'_{tR}}    &{{\bf A}'_{tv}}    &{{\bf A}'_{t\omega}}   &{{\bf A}'_{ta}}   \\
     {{{\bf A}'_{Rt}}}                &{{\bf A}'_{RR}}    &{{\bf A}'_{Rv}}    &{{\bf A}'_{R\omega}}    &{{\bf A}'_{Ra}}\\
     {{{\bf A}'_{vt}}}                &{{{\bf A}'_{vR}}}  &{{\bf A}'_{vv}}    &{{\bf A}'_{v\omega}}    &{{\bf A}'_{va}}\\
     {{\bf A}'_{\omega t}}            &{{\bf A}'_{\omega R}}                 &{{{\bf A}'_{\omega v}}}                 &{{\bf A}'_{\omega\omega}}&{{\bf A}'_{\omega a}}\\
     {{\bf A}'_{at}}                &{{\bf A}'_{aR}}                 &{{\bf A}'_{av}}                 &{{\bf A}'_{a\omega}}                         &{{\bf A}'_{aa}}
\end{array}}\right] \label{Eq: simplified Hession}
\end{equation}

According to the definition of the Hessian matrix for the Gaussian process Kalman filter (\ref{Eq: simplified Hession}), the LiDAR degeneracy metric \cite{degeneracy-zhangji} for optimization-based estimation can be adopted to the proposed filter-based estimation, which evaluates LiDAR degeneracy through the minimum eigenvalue of the Hessian matrix.
Without loss of generality, only the ${{\bf A}'_{tt}}$ and ${{\bf A}'_{RR}}$ are used in degeneracy analysis, because all elements in the state can be expressed by derivatives (from 0 to 2 order) of translation and rotation.
${{\bf A}'_{tt}}$ and ${{\bf A}'_{RR}}$ can be obtained by substituting (\ref{Eq: Jaco}) into (\ref{Eq: simplified Hession}).
\begin{equation}
% \small{
\begin{aligned}
&{{\bf A}'_{tt}} = \sum\limits_{i = 1}^{M} {{\bf{n}}_j^m} {\bf{n}}_j^{m,\top}\\
&{{\bf A}'_{RR}} =  - \sum\limits_{i = 1}^{M} {{{\left\lfloor {{{{\bf{P}}}_i}} \right\rfloor }_ \times }{{\bf{\check R}}_{t_k}^\top}{\bf{n}}_j^m} {\bf{n}}_j^{m,\top}{\bf{\check R}}_{t_k}{\left\lfloor {{{{\bf{P}}}_i}} \right\rfloor _ \times } \label{Eq: A'}
\end{aligned}
% }
\end{equation}
where ${{{\bf{P}}}_i} = \exp \left( {{{\left\lfloor {{\bm{\check\omega}}_{t_k}\Delta t} \right\rfloor }_ \times }} \right){{\bf{p}}_i} + \frac{1}{2}{\bf{\check a}}_{t_k}\Delta {t^2}$. 
The translational and rotational contribution for each point ${\bf p}_i$ is highly dependent on ${{\bf{n}}_j^m}$ and ${{{\left\lfloor {{{{\bf{P}}}_i}} \right\rfloor }_ \times }{{\bf{\check R}}_{t_k}^\top}{\bf{n}}_j^m}$, respectively.
Hence, the translational and rotational information matrices, ${\bf N}_{tt} \in \mathbb{R}^{M\times3}$ and ${\bf N}_{RR} \in \mathbb{R}^{M\times3}$, can be defined as:
\begin{equation}
% \small{
\begin{aligned}
&{{\bf{N}}_{tt}} = {\left[ { \ldots ,{\bf{n}}_j^m, \ldots } \right]^\top}\\
&{{\bf{N}}_{RR}} = {\left[ { \ldots ,\frac{{{{\left\lfloor {{{{\bf{P}}}_i}} \right\rfloor }_ \times }{{\bf{\check R}}_{t_k}^\top}{\bf{n}}_j^m}}{{{{\left\| {{{\left\lfloor {{{{\bf{P}}}_i}} \right\rfloor }_ \times }{{\bf{\check R}}_{t_k}^\top}{\bf{n}}_j^m} \right\|}^2}}}, \ldots } \right]^\top}
\end{aligned}\label{Eq: information matrix}
% }
\end{equation}
Inspired by \cite{xicp}, the translational and rotational contribution for a point cloud $\mathbb P$ can be quantified by projecting the information matrices onto the eigenspace of the translation Hessian ${{\bf A}_{tt}'}$ and rotation Hessian ${{\bf A}_{RR}'}$.
\begin{equation}
{{\bf{C}}_{tt}} = \left| {{{\bf{N}}_{tt}}{{\bf{V}}_{tt}}} \right|,{{\bf{C}}_{RR}} = \left| {{{\bf{N}}_{RR}}{{\bf{V}}_{RR}}} \right| \label{Eq: Contributio matrix}
\end{equation}
where $\left|\cdot\right|$ indicates the element-wise absolute value of the matrix, ${{\bf{V}}_{tt}} \in \mathbb{R}^{3\times3}$ and ${{\bf{V}}_{RR}}\in \mathbb{R}^{3\times3}$ are obtained through the eigen-analysis over ${{\bf A}'_{tt}}$ and ${{\bf A}'_{RR}}$, and ${{\bf{C}}_{tt}}, {{\bf{C}}_{RR}}\in\mathbb{R}^{M\times3}$ are the translational and rotational contribution matrixes, respectively. The $i$-th row of ${{\bf{C}}_{tt}}, {{\bf{C}}_{RR}}$ corresponds to the localizability contribution provided by ${\mathbf{p}_i\in\mathbb{P}}$.
To quantify the contribution of 6-DOF, six information pairs $\mathcal{L}_d(i) = \{i, {\bf C}(i,d)\}, d=1,\ldots,6$ are defined for each point ${\bf p}_i$, where ${\bf C}=\left[{\bf C}_{tt}, {\bf C}_{RR}\right]$. 
Hence, according to the sorting of the contributions for each point $\mathbf{p}_i$ to each DOF $d$, points are added to the selected point set $\mathbb{P}_s$ in descending order until the contributions of the points in the selected point set $\mathbb{P}_s$ satisfy the constraints defined in (\ref{Eq: score}) for all 6-DOF.
\begin{equation}
{S_d} > s,{S_d} = \sum\limits_{{{\bf{p}}_i} \in {{\mathbb{P}}_s}} {{\bf{C}}(i,d)}, d=1,\ldots,6 \label{Eq: score}
\end{equation}
where $s$ is the localizability threshold. 

Considering the definition of the contribution matrix (\ref{Eq: Contributio matrix}) relies on the point-to-voxel correspondence information, which process is expensive, especially when repeatedly computed within iterations. 
To tackle this problem, a normal estimation algorithm\footnote{\url{https://pcl.readthedocs.io/projects/tutorials/en/latest/normal_estimation.html}} is adopted to approximate the normal vector $\mathbf{n}_j^m \approx {\mathbf{\check R}_{t_k}}{{\bf{n}}_i}$ used in (\ref{Eq: A'}) and (\ref{Eq: information matrix}). Hence, (\ref{Eq: A'}) and (\ref{Eq: information matrix}) can be rewritten as:
\begin{equation}
\begin{array}{l}
{{\bf{A}}^\prime }_{tt} \approx  {\mathbf{\check R}_{t_k}}\sum\limits_{i = 1}^{M} {{{\bf{n}}_i}{\bf{n}}_i^ \top } { {\mathbf{\check R}_{t_k}}^ \top }\\
{{\bf{A}}^\prime }_{RR} \approx  - \sum\limits_{i = 1}^{M} {{{\left\lfloor {{{{\bf{P}}}_i}} \right\rfloor }_ \times }{{\bf{n}}_i}{\bf{n}}_i^ \top } {\left\lfloor {{{{\bf{P}}}_i}} \right\rfloor _ \times }\\
{{\bf{N}}_{tt}} \approx {\left[ { \ldots , {\mathbf{\check R}_{t_k}}{{\bf{n}}_i}, \ldots } \right]^ \top },{{\bf{N}}_{RR}} \approx {\left[ { \ldots ,\frac{{{{\left\lfloor {{{{\bf{P}}}_i}} \right\rfloor }_ \times }{{\bf{n}}_i}}}{{{{\left\| {{{\left\lfloor {{{{\bf{P}}}_i}} \right\rfloor }_ \times }{{\bf{n}}_i}} \right\|}^2}}}, \ldots } \right]^ \top }
\end{array} \label{Eq: approximated information}
\end{equation}
which process does not rely on the point-to-voxel correspondence information. The process of the localizability-aware point cloud sampling is given in Algorithm \ref{Alg: Localizability-aware point cloud sampling}.
\begin{algorithm}[t]
% \small{
    \caption{Localizability-aware sampling}
    \label{Alg: Localizability-aware point cloud sampling}
    \renewcommand{\thealgocf}{}
    \SetKwInOut{Input}{Input}
    \SetKwInOut{Output}{Output}
    \SetKwInOut{Begin}{Begin}
    \SetKwProg{Fn}{Function}{}\\		
        \Input{$\mathbb{P}$: Newly obtained point cloud,\\ $s$: Localizability threshold.}
        \Output{$\mathbb{P}_s$: The selected point cloud.}
    
    % \For{$i\gets 1 \,\, \textbf{to} \,\, M$}
    \For {each $\mathbf{p}_i\in \mathbb{P}$}
    {
        ${\bf n}_i \gets NormalEstimation({\bf p}_i)$\\
        Update $\mathbf{A}_{tt}',\mathbf{A}_{RR}',\mathbf{N}_{tt},\mathbf{N}_{RR}$ by substituting ${\bf n}_i$ into (\ref{Eq: approximated information})\\
    }
    Singular Value Decomposition (SVD): $\mathbf{A}_{tt}' = \mathbf{V}_{tt}\bm{\Lambda}_{tt}\mathbf{V}^\top_{tt}, \mathbf{A}_{RR}' = \mathbf{V}_{RR}\bm{\Lambda}_{RR}\mathbf{V}^\top_{RR}$\\
    Get the contribution matrix ${\bf C}=\left[{\bf C}_{tt}, {\bf C}_{RR}\right]$ by substituting $\mathbf{N}_{tt},\mathbf{N}_{RR}, \mathbf{V}_{tt},\mathbf{V}_{RR}$ into (\ref{Eq: Contributio matrix})\\
    \For{$d\gets 1 \,\, \textbf{to} \,\, 6$}
    {
        \tcp{$Card(\cdot)$ returns the cardinal number of a set}
        \For{$i\gets 1 \,\, \textbf{to} \,\, Card(\mathbb{P})$}
        {            
            Build information pair $\mathcal{L}_d(i)=\left\{i,\mathbf{C}(i,d)\right\}$\\
        }
        Sort $\mathcal{L}_d$ according to $\mathbf{C}(\cdots,d)$\\
        \For {$i\gets 1 \,\, \textbf{to} \,\, Card(\mathbb{P})$}
        {
            $S_d = S_d + \mathcal{L}_d(i).second$\\
            \If {$\mathbf{p}_i \notin \mathbb{P}_s$}
            {
                Add $\mathbf{p}_i$ into $\mathbb{P}_s$\\
            }
            \If {$S_d>s$}
            {
             Break\\
            }
        }
    }
    \Return {$\mathbb{P}_s$}
    % }    
\end{algorithm}
\begin{rem}
\textit{(Localizability-aware sampling) Based on the derivation on the Hessian matrix (from (\ref{Eq: p2v registration}) to (\ref{Eq: simplified Hession})), we expand the LiDAR degeneracy metric \cite{degeneracy-zhangji, xicp} for optimization-based estimation to the analysis of filter-based LiDAR odometry and quantify the localizability contribution for each point $\mathbf{p}_i\in \mathbb{P}$ in (\ref{Eq: A'})-(\ref{Eq: Contributio matrix}).
Thanks to above derivations, the proposed point cloud sampling strategy has the capability to modify the point number online with the consideration of localizability, so-called localizability-aware point cloud sampling.
In \cite{9561262, xicp}, the point sampling is developed with the prior knowledge of the point-to-map matching information, which process is time-consuming due to the LiDAR matching having to be performed for each iteration.
To tackle this problem, in this paper, a novel point sampling strategy is proposed without knowing the correspondences beforehand while considering the localizability.}
\end{rem}

\section{Voxel Map Management} \label{Sec: Voxel Map Management}
An adaptive voxelization method \cite{balm,voxelmap} is adopted to maintain the voxel map $\mathbb{M}$ for CTE-MLO with a hash data structure.
Each voxel $\mathbf{m}_j \in \mathbb{M}$ is clustered from a set of points $\mathbf{p}_i (i=1,2,\ldots,M_j)$ sampled at different times but belongs to the same plane in three-dimensional spatial space, which is indexed in a hash table.
With the Gaussian distribution assumption, a voxel $\mathbf{m}_j \sim \mathcal{N}(\bm{\mu}_j^m, \bm{\sigma}_j^m)$ can be described using a mean vector and a covariance matrix.
\begin{equation}
% \small{
\begin{aligned}
&{\bm{\mu }}_j^m = \frac{1}{M_j}\sum\limits_{i = 1}^{M_j} {{^W{\bf{p}}_i}}\\ 
&\begin{aligned}{\bm{\sigma }}_j^m &= \frac{1}{{M_j}}\sum\limits_{i = 1}^{M_j} {\left( {{^W{\bf{p}}_i} - {\bm{\mu }}_j^m} \right)} {\left( {{^W{\bf{p}}_i} - {\bm{\mu }}_j^m} \right)^\top}\\&=\frac{1}{M_j}\sum\limits_{i=1}^{M_j}{^W\mathbf{p}_i ^W\mathbf{p}_i^\top} - {{\bm \mu}_j^m}{\bm \mu}_j^{m,\top}\end{aligned}
\end{aligned}
% }
\end{equation}
Hence, each voxel in the voxel map is represented in a point-free formulation by a centroid ${\bm{\mu }}_j^m$, a covariance matrix ${\bm{\sigma}_j^m}$, and the number of points $M_j$ that contribute to cluster the voxel $\mathbf{m}_j$.
The voxel map stores all voxels in a hash table, and the hash key of each voxel is calculated as follows:
\begin{equation}
{K_j} = H\left( {\left\lfloor {\frac{{\bm{\mu} _j^m}}{r}} \right\rfloor } \right)
\end{equation}
where $H(\cdot)$ is a function that calculates the hash index $K_j\in \mathbb{R}$ using the centroid of the voxel ${\bm{\mu }}_j^m\in\mathbb{R}^3$, $r$ is the resolution of the voxel map, and $\left\lfloor {\cdot} \right\rfloor$ denotes the floor operation.

For a newly obtained point $\mathbf{p}_i (t_i\in[t_k, t_{k+1}))$, we first transform it into the inertial frame $^{W}{\mathbf{p}}_i=\mathbf{\bar R}(t_i)\mathbf{p}_i+\mathbf{\bar t}(t_i)$. The corresponding rotation $\mathbf{R}(t_i)$ and translation $\mathbf{t}(t_i)$ are estimated through the Gaussian process continuous-time LiDAR odometry (Section \ref{Sec: Continuous-Time LiDAR Odometry}).
Then, the corresponding voxel of $\mathbf{p}_i$ can be associated with the hash key $K_i = H\left( {\left\lfloor {\frac{{^W\mathbf{p}_i}}{r}} \right\rfloor } \right)$.
Inspired by \cite{BALM2}, the corresponding voxel can be updated incrementally.
\begin{equation}
\begin{aligned}
&{\bm{\mu}}_{j,\text{new}}^m = \frac{1}{{{M_{j,\text{new}}}}}\left( {M_j{\bm{\mu }}_j^m + \sum\limits_{{{\bf{p}}_i} \in {{\mathbb{P}}_j}} {{}^W{{\bf{p}}_i}} } \right)\\
&\begin{aligned}{\bm{\sigma }}_{j,\text{new}}^m &= \frac{1}{{{M_{j,\text{new}}}}}\left( {M_j\left( {{\bm{\sigma }}_j^m + {\bm{\mu }}_j^m{\bm{\mu}}_j^{m, \top }} \right) + \sum\limits_{{{\bf{p}}_i} \in {{\mathbb{P}}_j}} {{}^W{\bf{p}}_i^W{\bf{p}}_i^ \top } } \right) \\&- {\bm{\mu }}_{j,\text{new}}^m{\bm{\mu }}_{j,\text{new}}^{m, \top }\end{aligned}
\end{aligned}
\end{equation}
where ${M_{j,\text{new}}} = M_j + \textit{Card}\left( {{{\mathbb{P}}_j}} \right)$, $\mathbb{P}_j\in\mathbb{P}$ is the newly obtained point cloud that corresponds to the $j$-th voxel, and ${\bm{\mu }}_{j,\text{new}}^m$, ${\bm{\sigma}}_{j,\text{new}}^m$, and $M_{\text{new}}$ denote the updated centroid, covariance matrix, and number of points belonging to the $j$-th voxel, respectively.

After the voxel map update, voxels can be divided into different shapes depending on the relationships between the eigenvalues of the covariance matrix ${\bm \sigma}^m_j$.
The SVD (\ref{Eq: SVD voxel}) is performed to select the plane voxel that satisfies $\lambda_2 \gg \lambda_3$, and the normal vector of a plane voxel is defined as ${\bf{n}}_j^m={\bf{V}}_3$.
\begin{equation}
{{\bm{\sigma }}^m_j} = \left[ {{\bf{V}}_1,{\bf{V}}_2,{\bf{V}}_3} \right]
% \text{diag}([\lambda _1, \lambda _2, \lambda _3])
\left[ {\begin{array}{*{20}{c}}
{\lambda _1}&{0}&{0}\\
{0}&{\lambda _2}&{0}\\
{0}&{0}&{\lambda _3}
\end{array}} \right]
{\left[ {{\bf{V}}_1,{\bf{V}}_2,{\bf{V}}_3} \right]^\top} \label{Eq: SVD voxel}
\end{equation}
where $\lambda_1$, $\lambda_2$, and $\lambda_3$ are eigenvalues of the covariance matrix ${\bm \sigma}_j^m$ in descending order. $\mathbf{V}_1$, $\mathbf{V}_2$, and $\mathbf{V}_3$ are eigenvectors that correspond to $\lambda_1$, $\lambda_2$, and $\lambda_3$, respectively.

\section{Experimental Evaluations and Validations}
In this section, the effectiveness of the proposed CTE-MLO is demonstrated through various scenarios, including public datasets and real-world experiments.
\begin{rem}    
    \textit{(Extendability) The continuous-time trajectory formulation grants CTE-MLO exceptional extendability, allowing each point in a point stream (within the continuous-time trajectory period) to independently query the trajectory without synchronizing all points from different LiDARs to an instantaneous time.
    To demonstrate the extendability of CTE-MLO, in this section, various experiments are conducted with different LiDAR configurations, including spinning LiDAR (Ouster OS1, Robosense Helios 16/32), non-repetitive scan LiDAR (Livox Mid 360), and solid-state LiDAR (Robosense M1, Robosense Bpearl, Realsense L515).}\label{rem: Extendability}
\end{rem}
\subsection{Quantitative Evaluation on the NTU VIRAL Datasets}
The proposed CTE-MLO is evaluated quantitatively on the NTU VIRAL dataset \cite{ntuviral}, which provides point clouds captured by two lightweight ouster OS1 LiDARs (deployed in horizontal form and vertical form respectively), and ground truth from the Leica laser tracker.
Each experiment is conducted on a computer equipped with an Intel Core i9-13900KF.
The comparative methods include SOTA MLO MLOAM \cite{MLOAM} and a continuous-time LiDAR odometry MARS\cite{MARS}.
For a fair comparison, we merge the LiDAR point from two OS1 LiDAR into a single LiDAR scan through the LiDAR synchronization package\footnote{\url{https://github.com/brytsknguyen/slict/blob/master/src/MergeLidar.cpp}} provided by \cite{SLICT2}, which enables MARS \cite{MARS} as a MLO.
It is worth noting that all methods are implemented without loop closure.
Moreover, an ablation study on the proposed method is performed to understand the influence of the number of LiDAR and localizability-aware point cloud sampling.
We implement the proposed algorithm in various configurations, including 1) CT-LO: a continuous-time single LiDAR odometry developed through the Kalman filter-based Gaussian process estimation (Section \ref{Sec: Continuous-Time LiDAR Odometry}). Horizontal LiDAR is used for CT-LO; 2) CT-MLO: a continuous-time MLO that utilizes LiDAR points from multiple LiDARs using the decentralized multi-LiDAR synchronization scheme (Section \ref{Sec: Decentralized Multi-LiDAR Synchronization}); 3) CTE-MLO: a continuous-time and efficient MLO with the localizability-aware point cloud sampling illustrated in Section \ref{sec: Localizability-aware Point Cloud Sampling}, which is the full algorithm proposed in this paper.
\subsubsection{Ablation Study} \label{sec: Ablation Study}
\begin{figure}[!t]
	\centering	
	\subfigure[Trajectories]{
		\includegraphics[width=\linewidth]{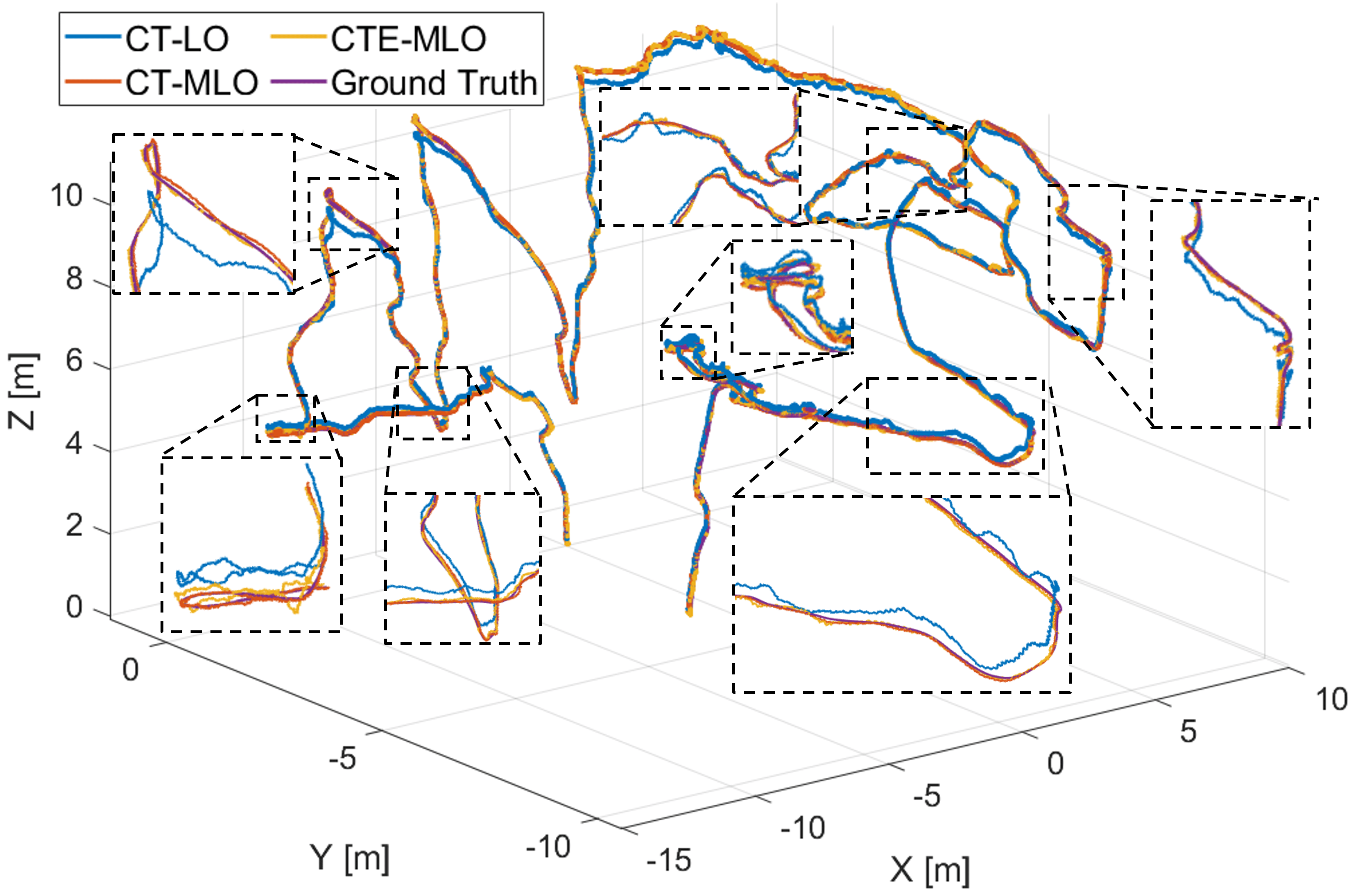}
		\label{Fig: tnp02_traj}		
	}
	% \quad
	\subfigure[Real-time performance]{
		\includegraphics[width=0.9\linewidth]{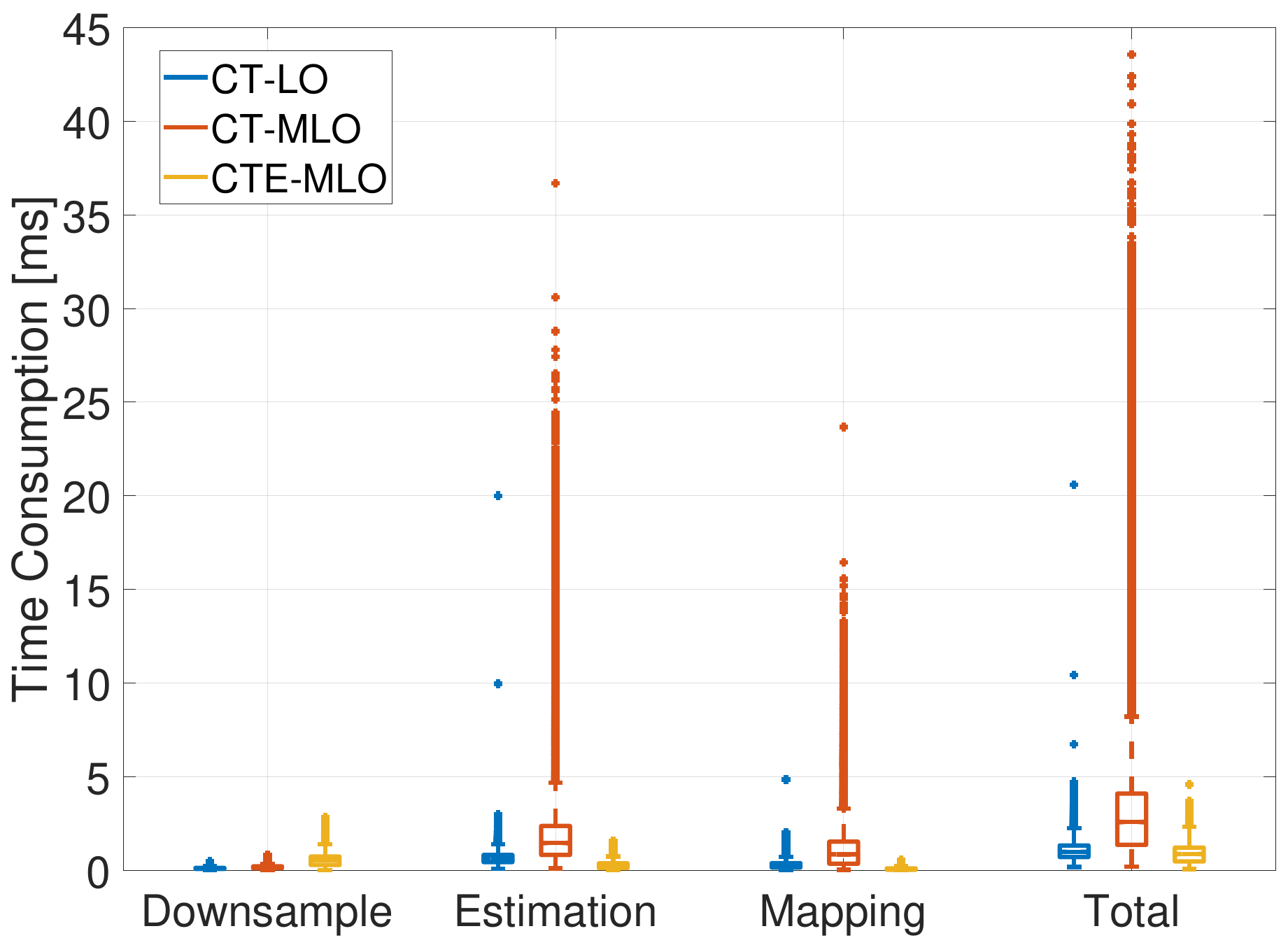} 
		\label{Fig: tnp02_time}
	}
	\caption{Benchmark results of CT-LO, CT-MLO, and CTE-MLO under the \textit{tnp02} sequence of the NTU VIRAL dataset.} \label{Fig: NTU_mapping}
    % \vspace{-1em}
\end{figure}
To illustrate the effectiveness of the multi-LiDAR fusion, an ablation study over single LiDAR odometry (CT-LO) and multi-LiDAR odometry (CT-MLO and CTE-MLO) is conducted on the NTU VIRAL dataset \cite{ntuviral}.
Localization for a MAV flying at a high altitude is challenging because the horizontal LiDAR can only provide observation around a MAV with very limited points belonging to the ground or ceiling.
As shown in Fig. \ref{Fig: tnp02_traj}, the scarcity of geometric constraint over the $z$-axis leads CT-LO can not provide accurate estimation for MAV when compared with the multi-LiDAR version of the proposed method called CT-MLO and CTE-MLO.
Thanks to the decentralized multi-LiDAR synchronization scheme introduced in Section \ref{Sec: Decentralized Multi-LiDAR Synchronization}, CT-MLO outperformed CT-LO in accuracy especially along the $z$-axis by fully leveraging the geometric information observed from both horizontal LiDAR and vertical LiDAR.
To quantitatively evaluate the accuracy of MLO, the Absolute Trajectory Error (ATE) results of each method over NTU VIRAL datasets are shown in Table \ref{tab: NTU dataset ATE}.
CT-MLO exhibits excellent accuracy on 17 out of 18 sequences of the NTU VIRAL dataset.
However, the large-size point cloud generated by multiple LiDARs worsens the computational complexity.
As shown in Fig. \ref{Fig: tnp02_time}, despite the average time consumption of CT-MLO being less than the time interval $\Delta t = 10ms$, the total time consumption of CT-MLO has a large number of outliers greater than $\Delta t$.
This leads to the failure of CT-MLO in sequence \textit{spms02} due to CT-MLO has to drop a lot of LiDAR scans to achieve real-time performance.
Hence, motivated by developing a real-time and accurate MLO, CTE-MLO is proposed by introducing the localizability-aware point cloud sampling strategy into CT-MLO.
As shown in Fig. \ref{Fig: tnp02_time}, the localizability-aware point cloud sampling strategy can significantly improve the estimation efficiency, and the time consumption of CTE-MLO is strictly less than the time interval $\Delta t$, which means CTE-MLO can achieve real-time without dropping any scan.
\begin{table*}[!t] \centering
    \setlength{\tabcolsep}{2.2pt} 
    \centering
	\caption{Absolute Trajectory Error (ATE, Meters) over NTU VIRAL dataset.}
	\label{tab: NTU dataset ATE}
	\begin{threeparttable}
    \begin{tabular}{l || c c c c c c c c c c c c c c c c c c||c}
    \hline\hline
    Method  &eee01 &eee02  &eee03 &nya01 &nya02  &nya03 &sbs01 &sbs02  &sbs03 &rtp01 &rtp02  &rtp03 &spms01 &spms02  &spms03 &tnp 01 &tnp02  &tnp03 &Average\\ \hline
    CT-LO &0.08 &0.07 &0.11 &0.05 &0.09 &0.11 &0.09 &0.08 &\color{blue} 0.08 &- &0.13 &0.15 &0.22 &\color{blue} 0.31 &- &- &0.12 &0.10 &0.12\\
    CT-MLO &\color{blue} 0.07 &\color{blue}0.07 &0.11 &\color{blue} 0.05 &0.08 &0.10 &\color{blue}0.09 &\color{blue}0.08 &0.08 &\color{blue}0.13 &0.13 &\color{blue} 0.14 &0.21 &- &0.20 &\color{blue}0.08 &\color{blue}0.07 &\color{blue}0.08 &\color{blue}0.10\\
    CTE-MLO &0.08 &0.07 &0.12 &0.06 &0.09 &0.10 &0.09 &0.08 &0.09 &0.13 &0.14 &0.14 &\color{blue}0.21 &0.33 &\color{blue} 0.20 &0.09 &0.09 &0.10 &0.12\\   
    MLOAM &0.23 &0.16 &0.23 &0.13 &0.19 &0.22 &0.18 &0.16 &0.17 &0.41 &0.34 &1.04 &0.79 &11.72 &6.11 &0.15 &0.11 &0.15 &1.25\\
    MARS &0.19 &0.08 &\color{blue}0.10 &0.06 &\color{blue}0.06 &\color{blue} 0.08 &0.13 &0.14 &0.20 &0.15 &\color{blue} 0.08 &0.14 &0.94 &- &- &0.08 &0.09 &0.11 &0.16\\
    \hline\hline
    \end{tabular}
\begin{tablenotes}
    \footnotesize
    \item[1] - denotes the system failure, and the best result is highlighted in {\color{blue}{Blue}}.
    % \item[3] Aver denotes the average ATE.
\end{tablenotes}
\end{threeparttable}
% \vspace{-1em}
\end{table*}

\begin{figure*}[!t]
	\centering    
	\subfigure[Localizability]{
		\includegraphics[width=0.49\linewidth]{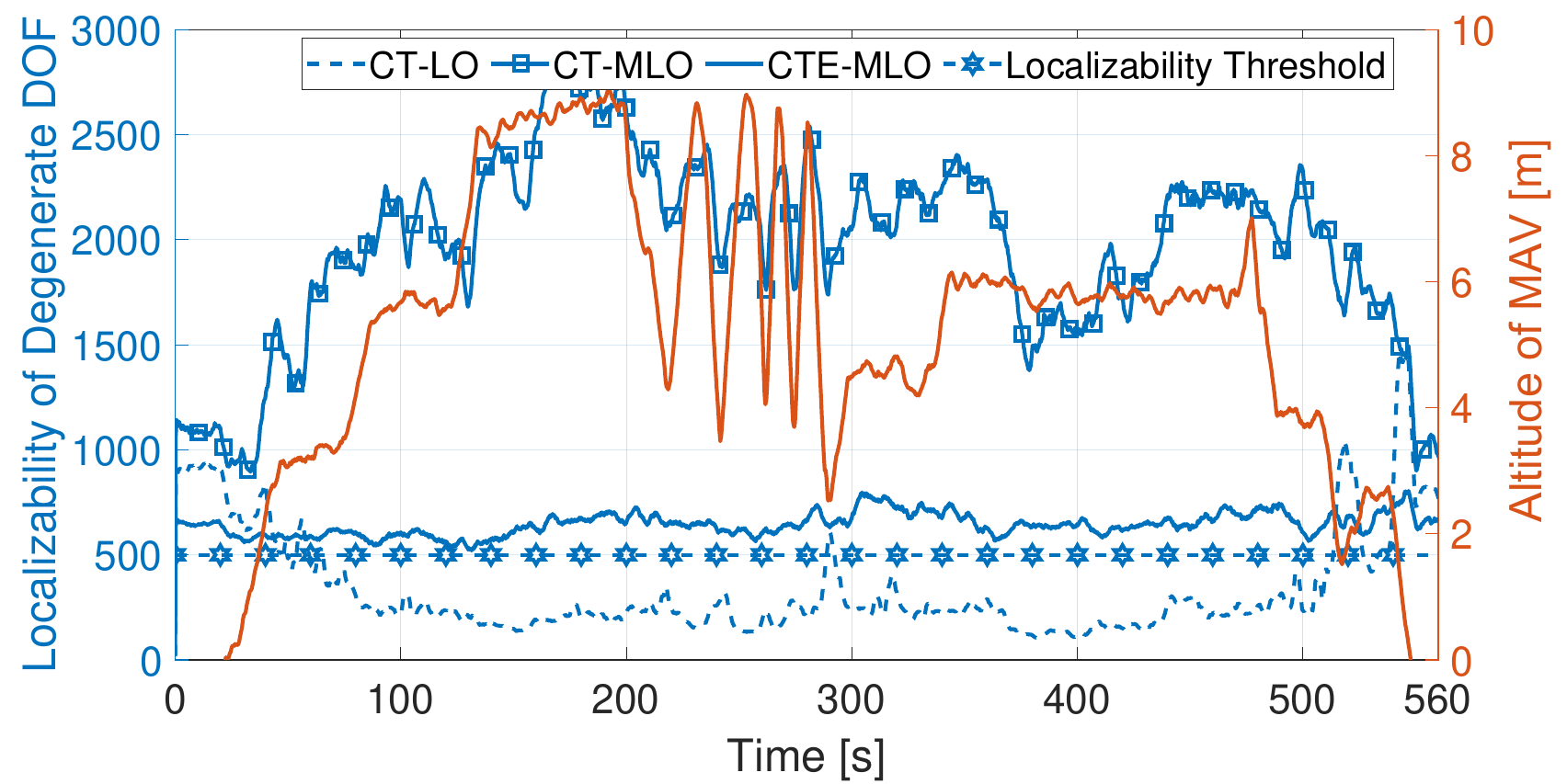}	
        \label{Fig: Scoreandheight}
	}
        % \vspace{-1em}
	% \quad
	\subfigure[Number of points used in estimation]{
		\includegraphics[width=0.45\linewidth]{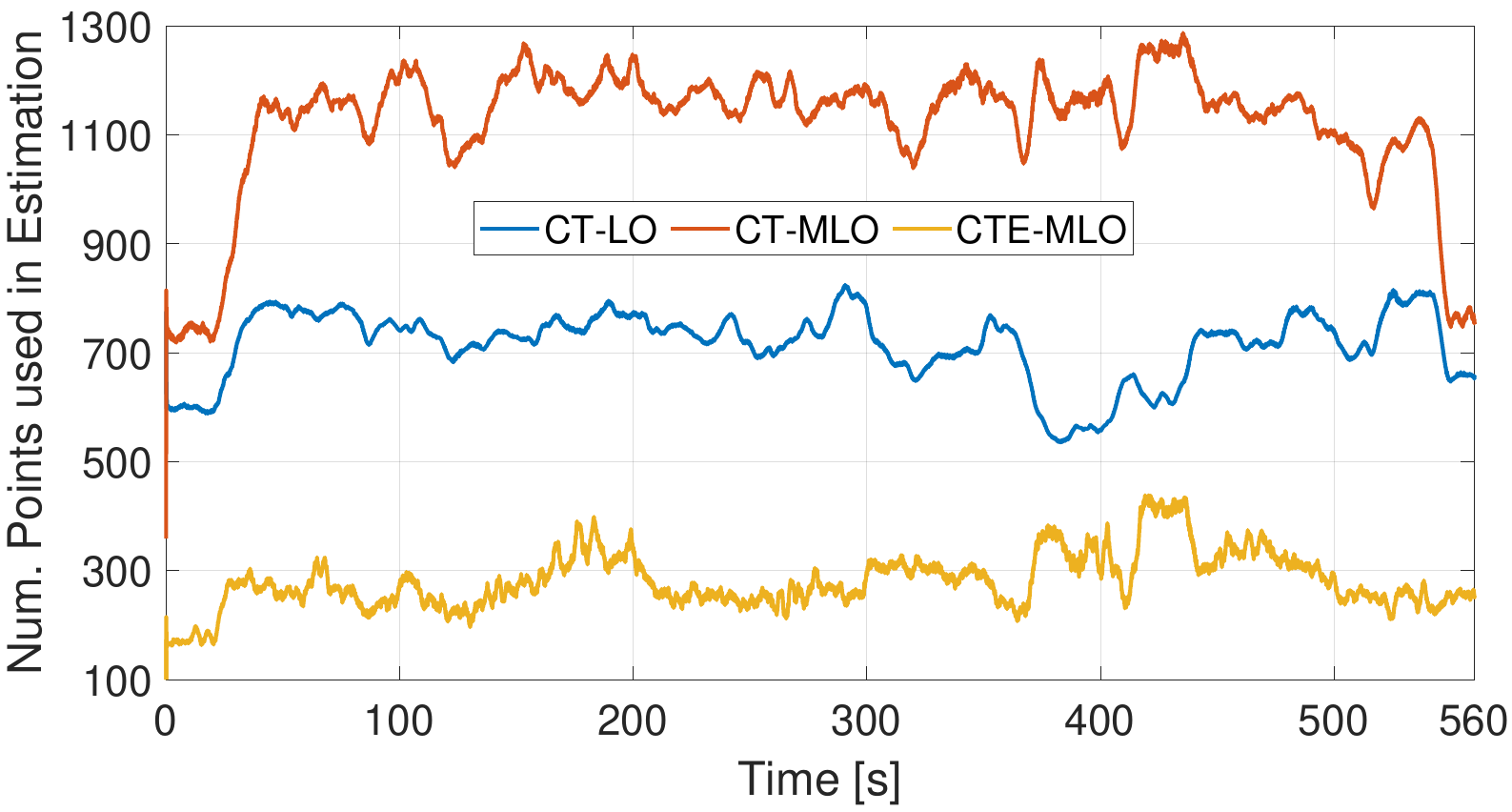} 
		\label{Fig: PointNum}
	}
    % \vspace{-1em}
        \subfigure[Point stream]{
		\includegraphics[width=0.95\linewidth]{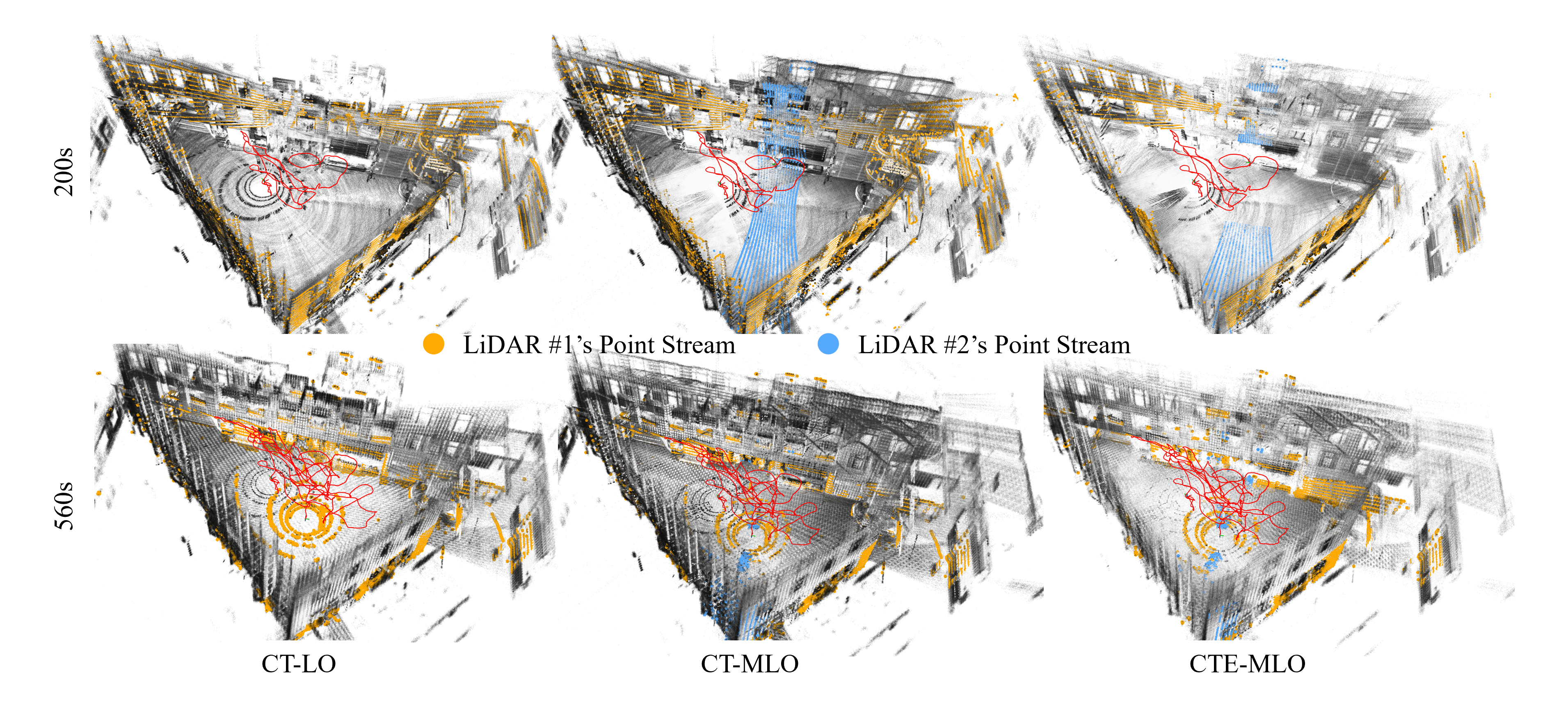} 
		\label{Fig: tnp01_degeneration_map}
	}
        % \vspace{-1em}
	\caption{Localizability and used points of CT-LO, CT-MLO, and CTE-MLO under a degenerate sequence (\textit{tnp01}) of the NTU VIRAL dataset.} \label{Fig: degeneration}
    % \vspace{-0.5em}
\end{figure*}

To justify the effectiveness of the localizability-aware point cloud sampling, as shown in Fig. \ref{Fig: degeneration}, the localizability and used points of CT-LO, CT-MLO, and CTE-MLO are compared under one of the degenerate sequences over the NTU-VIRAL dataset.
For visualization, the point stream shown in Fig. \ref{Fig: tnp01_degeneration_map} comprises points accumulated within $0.1s$.
From the top row of Fig. \ref{Fig: tnp01_degeneration_map}, there are almost no points belonging to the ground or ceiling to provide $z$-axis constraints for CT-LO when the MAV is flying at a relatively high altitude, which is attributed to the limited vertical FOV of the horizontal LiDAR.
CT-MLO prevents the degeneration of CT-LO by leveraging the merged point stream observations from horizontal LiDAR and vertical LiDAR.
As shown in Fig. \ref{Fig: Scoreandheight}, although CT-MLO ensures the localizability using the observation from multiple LiDAR, the dense point cloud preserves a high level of geometric constraint redundancy (the localizability of CT-MLO is much greater than the empirical localizability threshold $s=500$).
This redundancy unnecessarily slows down the CT-MLO and may cause estimation failure under real-time constraints.
Thanks to the localizability-aware point cloud sampling strategy, which quantifies the localizability contribution for each point, CTE-MLO can ensure localizability using a minimal number of points with high localizability contributions.
This enables CTE-MLO to avoid degeneration by using fewer points compared to CT-LO and CT-MLO (Fig. \ref{Fig: PointNum}).
As shown in Fig. \ref{Fig: Scoreandheight} and Fig. \ref{Fig: tnp01_degeneration_map}, the localizability-aware point cloud sampling strategy defined in Section \ref{sec: Localizability-aware Point Cloud Sampling} can effectively evaluate the localizability of LiDAR odometry.
When the MAV is flying at a relatively high altitude, the proposed localizability-aware point cloud sampling strategy can detect the degeneration of CT-LO.
When the MAV flies at low altitudes, CT-LO and CT-MLO exhibit similar localizability because the horizontal LiDAR can provide sufficient ground observations (as shown in the bottom row of Fig. \ref{Fig: tnp01_degeneration_map}).
Thanks to the integration of tightly coupled multi-LiDAR estimation and the efficient localizability-aware point cloud sampling strategy, CTE-MLO exhibits excellent localization and mapping consistency across the 18 data sequences and reaches a $54\%$ estimation efficiency improvement when compared with the CT-MLO (as shown in Table \ref{tab: NTU dataset time}).
\begin{table*}[!t] \centering
    \belowrulesep=0pt
    \aboverulesep=0pt
    % \scriptsize
    \setlength{\tabcolsep}{4.2pt}
    \centering
    \begin{threeparttable}
    \caption{Efficiency for processing one scan over the NTU VIRAL dataset.}
    \label{tab: NTU dataset time}
    \begin{tabular}{l||c c c c c c c c c c c c c c c c c c c c}
    \hline\hline
    \multirow{2}{*}{}    					&\multicolumn{4}{c}{CT-LO} 	&\multicolumn{4}{c}{CT-MLO} 	&\multicolumn{4}{c}{CTE-MLO} 	&\multicolumn{4}{c}{MLOAM} 	&\multicolumn{4}{c}{MARS} \\ \cmidrule(l){2-5}\cmidrule(l){6-9}\cmidrule(l){10-13}\cmidrule(l){14-17} \cmidrule(l){18-21}
    &Total &DS &ES &Map &Total &DS &ES &Map &Total &DS &ES &Map &Total &DS &ES &Map &Total &DS &ES &Map\\\hline
    eee01 &0.17 &0.01 &0.11 &0.06 &0.42 &0.01 &0.25 &0.16 &{\color{blue}0.12} &0.06 &{\color{blue}0.04} &{\color{blue}0.02} &1.56 &0.45 &0.62 &0.49 &0.21 &{\color{blue}0.00} &0.16 &0.04\\
    eee02 &0.12 &0.01 &0.08 &0.03 &0.17 &0.01 &0.11 &0.05 &{\color{blue}0.11} &0.06 &{\color{blue}0.03} &{\color{blue}0.01} &1.04 &0.26 &0.55 &0.23 &0.18 &{\color{blue}0.00} &0.15 &0.03\\
    eee03 &0.10 &0.01 &0.07 &0.03 &0.15 &0.01 &0.10 &0.04 &{\color{blue}0.10} &0.06 &{\color{blue}0.03} &{\color{blue}0.01} &1.12 &0.38 &0.53 &0.21 &0.20 &{\color{blue}0.00} &0.15 &0.04\\
    nya01 &0.16 &0.01 &0.10 &0.05 &0.11 &0.01 &0.07 &0.02 &{\color{blue}0.09} &0.05 &{\color{blue}0.03} &{\color{blue}0.01} &0.94 &0.31 &0.49 &0.12 &0.17 &{\color{blue}0.00} &0.15 &0.02\\
    nya02 &{\color{blue}0.07} &0.01 &0.05 &0.01 &0.11 &0.01 &0.08 &0.03 &0.09 &0.05 &{\color{blue}0.03} &{\color{blue}0.01} &0.90 &0.26 &0.51 &0.12 &0.18 &{\color{blue}0.00} &0.16 &0.02\\
    nya03 &{\color{blue}0.07} &0.01 &0.05 &0.02 &0.33 &0.01 &0.21 &0.11 &0.12 &0.06 &{\color{blue}0.04} &{\color{blue}0.01} &1.09 &0.30 &0.51 &0.28 &0.19 &{\color{blue}0.00} &0.15 &0.03\\
    sbs01 &{\color{blue}0.09} &0.01 &0.05 &0.02 &0.14 &0.01 &0.09 &0.04 &0.09 &0.05 &{\color{blue}0.03} &{\color{blue}0.01} &0.92 &0.26 &0.45 &0.21 &0.16 &{\color{blue}0.00} &0.13 &0.02\\
    sbs02 &{\color{blue}0.09} &0.01 &0.06 &0.02 &0.12 &0.01 &0.08 &0.03 &0.10 &0.06 &{\color{blue}0.03} &{\color{blue}0.01} &1.13 &0.41 &0.47 &0.24 &0.15 &{\color{blue}0.00} &0.13 &0.02\\
    sbs03 &0.14 &0.01 &0.09 &0.04 &0.19 &0.01 &0.12 &0.06 &{\color{blue}0.10} &0.06 &{\color{blue}0.03} &{\color{blue}0.01} &1.10 &0.37 &0.47 &0.25 &0.16 &{\color{blue}0.00} &0.13 &0.02\\
    rtp01 &- &- &- &- &0.31 &0.02 &0.18 &0.11 &{\color{blue}0.14} &0.07 &{\color{blue}0.05} &{\color{blue}0.02} &1.89 &0.82 &0.54 &0.53 &0.27 &{\color{blue}0.00} &0.22 &0.05\\
    rtp02 &0.13 &0.01 &0.08 &0.04 &0.19 &0.02 &0.12 &0.05 &{\color{blue}0.13} &0.06 &{\color{blue}0.04} &{\color{blue}0.02} &1.38 &0.51 &0.50 &0.37 &0.26 &{\color{blue}0.00} &0.22 &0.03\\
    rtp03 &{\color{blue}0.07} &0.01 &{\color{blue}0.04} &{\color{blue}0.01} &0.21 &0.02 &0.13 &0.06 &0.13 &0.07 &0.04 &0.02 &1.43 &0.55 &0.50 &0.38 &0.26 &{\color{blue}0.00} &0.20 &0.04\\
    spms01 &0.23 &0.01 &0.13 &0.08 &0.36 &0.01 &0.22 &0.12 &{\color{blue}0.17} &0.08 &{\color{blue}0.06} &{\color{blue}0.03} &1.77 &0.61 &0.52 &0.64 &0.28 &{\color{blue}0.00} &0.20 &0.08\\
    spms02 &0.28 &{\color{blue}0.00} &0.18 &0.10 &- &- &- &- &{\color{blue}0.18} &0.07 &{\color{blue}0.07} &{\color{blue}0.04} &1.43 &0.36 &0.42 &0.64 &- &- &- &-\\
    spms03 &- &- &- &- &0.37 &{\color{blue}0.01} &0.22 &0.13 &{\color{blue}0.14} &0.07 &{\color{blue}0.04} &{\color{blue}0.02} &1.24 &0.35 &0.49 &0.39 &- &- &- &-\\
    tnp01 &- &- &- &- &0.11 &0.02 &0.07 &0.02 &{\color{blue}0.09} &0.05 &{\color{blue}0.03} &{\color{blue}0.01} &0.77 &0.25 &0.39 &0.12 &0.20 &{\color{blue}0.00} &0.18 &0.01\\
    tnp02 &{\color{blue}0.10} &0.01 &0.07 &0.02 &0.37 &0.02 &0.21 &0.14 &0.11 &0.06 &{\color{blue}0.03} &{\color{blue}0.01} &0.74 &0.12 &0.40 &0.21 &0.19 &{\color{blue}0.00} &0.17 &0.01\\
    tnp03 &0.15 &0.01 &0.09 &0.05 &0.74 &0.02 &0.44 &0.29 &{\color{blue}0.10} &0.06 &{\color{blue}0.03} &{\color{blue}0.01} &0.65 &0.10 &0.38 &0.16 &0.19 &{\color{blue}0.00} &0.17 &0.01\\
    \hline
    Average &0.13 &0.01 &0.08 &0.04 &0.26 &0.01 &0.16 &0.09 &{\color{blue}0.12} &0.06 &{\color{blue}0.04} &{\color{blue}0.02} &1.17 &0.37 &0.49 &0.31 &0.20 &{\color{blue}0.00} &0.17 &0.03\\
    \hline\hline
    \end{tabular}
\begin{tablenotes}
    \footnotesize
    \item[1] - denotes the system failure, and the best result is highlighted in {\color{blue}{Blue}}.
    % \item[3] Aver denotes the average efficiency.
\end{tablenotes}
\end{threeparttable}
\vspace{-1.0em}
\end{table*}
\subsubsection{Accuracy evaluation} 
\begin{figure*}[!t]\centering
\includegraphics[width=\linewidth]{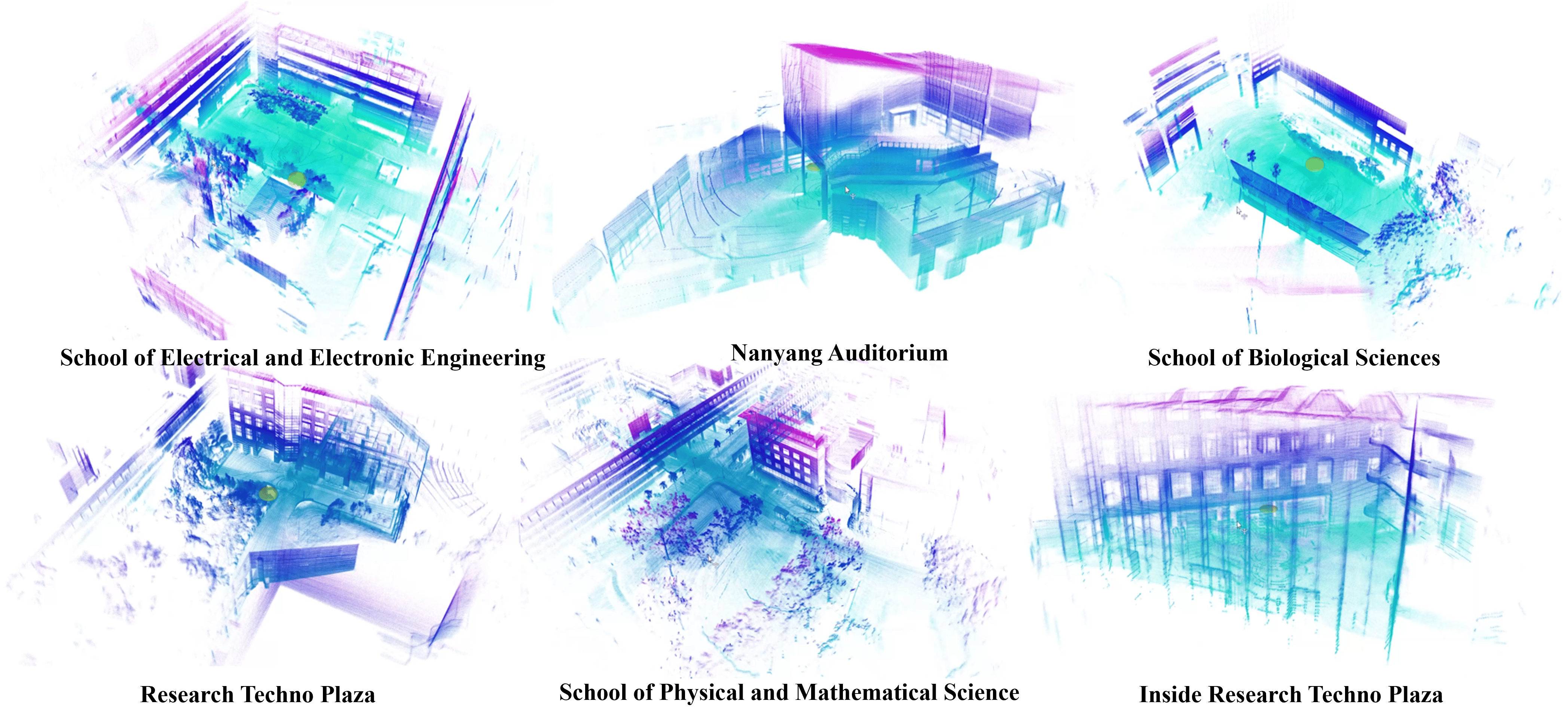}
% \vspace{-2em}
\caption{Reconstruction result of a part of NTU landmarks using the proposed CTE-MLO on the NTU VIRAL dataset.}\label{Fig: MAPPING NTU VIRAL}
% \vspace{-1em}
\end{figure*}

Table \ref{tab: NTU dataset ATE} summarized the ATE of the proposed method and SOTA LiDAR odometry, including the SOTA MLO MLOAM\cite{MLOAM} and multi-LiDAR-augmented continuous-time LiDAR odometry MARS\cite{MARS}, over the NTU VIRAL dataset\cite{ntuviral}.
The average accuracy of CTE-MLO outperforms both MLOAM \cite{MLOAM} and MARS \cite{MARS}.
Thanks to the Gaussian process Kalman filter derived in Section \ref{Sec: Continuous-Time LiDAR Odometry}, which considers the continuous-time sampling characteristics of LiDAR in a Kalman filter, CTE-MLO can estimate the trajectory using point stream from multiple LiDARs 
without estimating the state of prime LiDAR using the point cloud undistortion technique like MLOAM \cite{MLOAM}.
When compared with MARS\cite{MARS}, CTE-MLO achieves continuous-time estimation using Gaussian process regression, which dispenses with the estimation of control points not belong to the current time domain.
Due to future control points estimation being inevitable for the B-spline-based continuous-time estimator MARS\cite{MARS}, a keyframe strategy is adopted in MARS\cite{MARS} to enable continuous-time trajectory estimation in real time.
Nonetheless, the keyframe strategy is a trade-off between accuracy and real-time performance, which also leads to failure during aggressive rotational and translational motions in \textit{spms02} and \textit{spms03}.
Benefited by Gaussian process Kalman filter and localizability-aware point cloud sampling scheme, CTE-MLO can provide consistent estimation over 18 sequences of NTU VIRAL dataset, and the average ATE of CTE-MLO achieves $0.122m$ which reaches a $25\%-90\%$ accuracy improvement over MARS \cite{MARS} and MLOAM \cite{MLOAM}.
The real-time reconstructed 3D map is shown in Fig. \ref{Fig: MAPPING NTU VIRAL}.
The high-quality point cloud demonstrates that the CTE-MLO is highly accurate and is able to provide high-precision reconstruction using LiDAR point cloud without any extra sensors, such as IMU, camera, and GNSS.
\subsubsection{Running time evaluation} \label{Sec: Running time evaluation}
Directly comparing the processing time of the proposed method with SOTA methods would be unfair due to the adoption of the LiDAR splitting technique in the multi-LiDAR synchronization, which technique significantly reduces the dimension of the Jacobian matrix used for Hessian construction.
Therefore, the decentralized multi-LiDAR synchronization scheme implicit a high-level compression of computational complexity in the state update process (\ref{Eq: kalman gain matrix})-(\ref{Eq: state update ekf}).
For a fair comparison,
the efficiency is measured by the ratio between processing time $t_{\text{process}}$
and the time interval ($\Delta t=10ms$ for CT-LO, CT-MLO, and CTE-MLO; $\Delta t=100ms$ for MLOAM\cite{MLOAM} and MARS\cite{MARS}).
\begin{equation}
    \text{Efficiency} = t_{\text{process}}/\Delta t\label{Eq: Efficiency}
\end{equation}
Table \ref{tab: NTU dataset time} shows the average efficiency for processing one scan of all the benchmark methods.
The efficiency of each component, including sampling (``DS" in Table \ref{tab: NTU dataset time}), estimation (``ES" in Table \ref{tab: NTU dataset time}), and mapping (``Map" in Table \ref{tab: NTU dataset time}), are compared across all 18 sequences.
Moreover, the total efficiency (``Total" in Table \ref{tab: NTU dataset time}) is also measured by the ratio between total processing time (gathered all the time consumption for processing one scan, including the point cloud sampling, estimation, and mapping) and the time interval $\Delta t$.
The localizability-aware point cloud sampling allows CTE-MLO to select points with high localizability contribution, which induces a high-level compression of the dense point cloud merged from multiple LiDARs.
As noted in Section \ref{sec: Ablation Study} (Ablation Study), the localizability-aware point cloud sampling enables CTE-MLO to achieve $75\%$ and $78\%$ efficiency improvements in estimation and mapping, respectively, compared to CT-MLO.
Besides the localizability-aware point cloud sampling, combining the Gaussian process regression with the Kalman filter can also improve the estimation efficiency of LiDAR odometry, which enables continuous-time estimation within only a few linear iterations.
To demonstrate this, the estimation efficiency of CT-MLO is compared with that of MLOAM\cite{MLOAM} and MARS\cite{MARS}, both of which utilize nonlinear solvers for estimation.
Compared to MLOAM\cite{MLOAM}, CT-MLO improves estimation efficiency by more than $67\%$.
% Compared with MLOAM\cite{MLOAM}, CT-MLO achieves more than $67\%$ of the efficiency improvements.
Even compared to MARS\cite{MARS}, using voxel-to-voxel registration, CT-MLO demonstrates better estimation efficiency with point-to-voxel registration.
Benefiting from the combination of the Gaussian process regression with the Kalman filter and the localizability-aware point cloud sampling, as results are shown in Table \ref{tab: NTU dataset time}, CTE-MLO achieves the best average estimation and mapping efficiency when compared with SOTA MLO over all 18 sequences.
CTE-MLO achieves $43\%-90\%$ efficiency improvement compared to MARS\cite{MARS} and MLOAM\cite{MLOAM}, respectively.
The average efficiency of CTE-MLO achieves $0.116$ which means the proposed method only requires $11.6\%$ of the time interval $\Delta t=10ms$ to realize localization and mapping.
\vspace{-0.5em}
\subsection{Real-world Applications}
\begin{figure*}[!t]
	\centering	
 	\subfigure[Truck]{
		\includegraphics[width=0.31\linewidth]{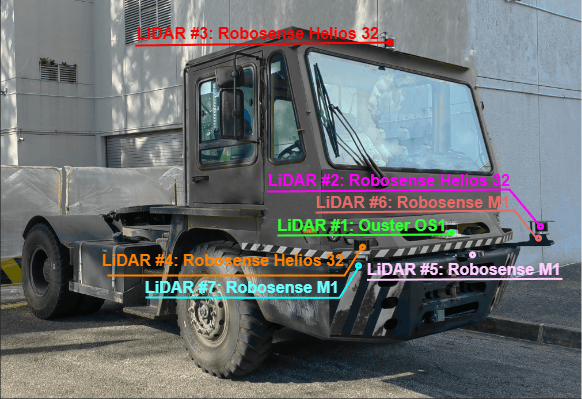} 
		\label{Fig: Prime Mover}
	}
	\quad
	\subfigure[Sweeper]{
		\includegraphics[width=0.28\linewidth]{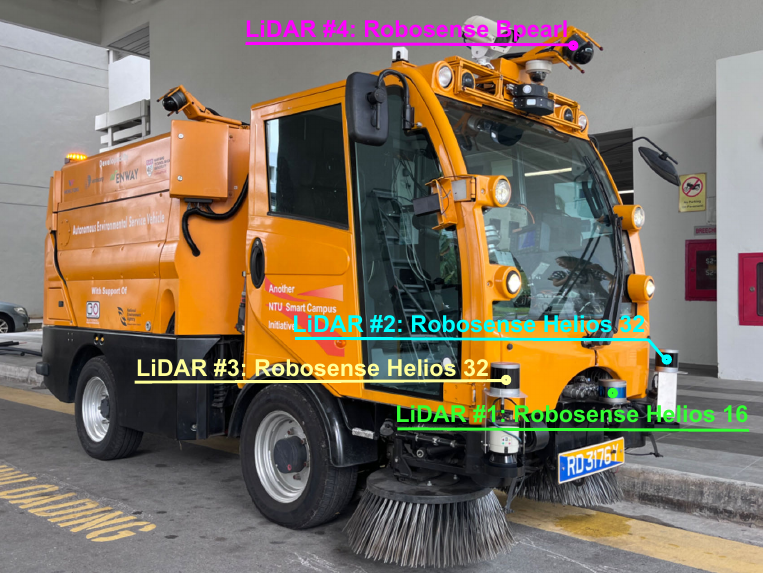}
		\label{Fig: sweeper}		
	}
        \quad
        \subfigure[MAV]{
		\includegraphics[width=0.31\linewidth]{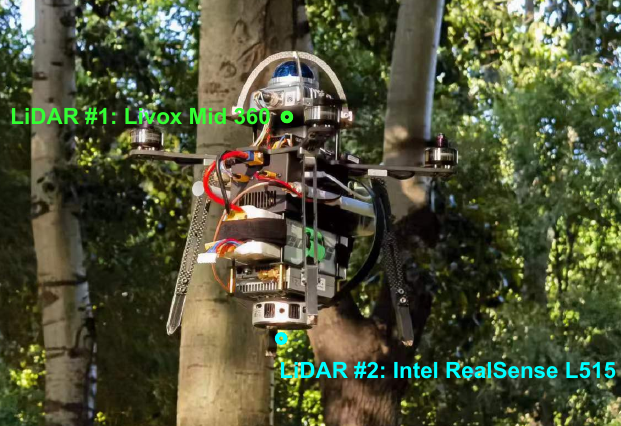}
		\label{Fig: MAV}		
	}
        % \vspace{-1em}
	\caption{Platforms used in real-word applications.} \label{Fig: platform}
        % \vspace{-1em}
\end{figure*}

To attest to the practicality, CTE-MLO is applied to enable estimation and mapping for three autonomous platforms, called truck, sweeper, and MAV.
As shown in Fig. \ref{Fig: platform}, both truck and sweeper are equipped with multiple heterogeneous LiDARs to enable panoramic LiDAR observation in horizontal view and onboard computer with an Intel Core i9-13900H CPU.
Different from the ground vehicles, which are not sensitive to the weight and power of onboard equipment, two light-weight solid-state LiDARs are mounted on MAV, and an edge computation platform (Nvidia Orin NX) is devoted as the onboard computer.
We conduct multiple validations of the proposed method in three scenarios, including seaport, campus, and field forest environments.
As the detail of real-world application sequences shown in Table \ref{Tab: Data sequences information}, the total validation duration reaches 3 hours, covering a total distance of over 44 kilometers.
\begin{table*}[!t] \centering
    \setlength{\tabcolsep}{3.5pt} 
    \centering
    \caption{Data sequences information used in real-world applications.}
	\label{Tab: Data sequences information}
	\begin{threeparttable}
    \begin{tabular}{l || c c c c c c}
    \hline\hline
    Scenario  &Sequence &Distance (m) &Druation (min:sec) &LiDARs &LiDAR Data Stream\\ \hline
    \multirow{2}{*}{Campus} &Campus 01 &966.0 &07:24 &3 ML+1 SSL&\usym{1F5F8}\\
                            &Campus 02 &4420.6&21:10 &3 ML+1 SSL&\usym{1F5F8}\\
                            &Campus 03 &5467.3&26:12 &3 ML+1 SSL&\usym{1F5F8}\\
    \multirow{4}{*}{Port}   &Port 01 &12148.4 &34:40 &4 ML+3 SSL &LiDAR $\#1$: 591-654s, 675-801s, 1269-1597s, 1609-1664s. LiDAR $\#7$: \usym{2717}\\
                            &Port 02 &9796.7 &44:48 &4 ML+3 SSL&LiDAR $\#7$: \usym{2717}\\
                            &Port 03 &11009.7 &44:57 &4 ML+3 SSL&LiDAR $\#5$: \usym{2717} LiDAR $\#7$: \usym{2717}\\
    Forest   &Forest 01 &526.0 &05:14 &2 SSL &\usym{1F5F8}\\                        
    \hline\hline
    \end{tabular}
\begin{tablenotes}
    \footnotesize
    \item[1] ML and SSL are denoting Mechanical LiDAR and Solid State LiDAR, respectively.
    \item[2] In sequence Port 01, LiDAR $\#1$ experienced data lose due to overly high temperatures.
    \item[3] \usym{1F5F8} denotes that all LiDARs equipped on the platform are working fine, and \usym{2717} denotes the data of LiDAR is totally lost due to the hardware failure.
\end{tablenotes}
\end{threeparttable}
% \vspace{-1em}
\end{table*}
\begin{figure*}[!t]\centering
\includegraphics[width=\linewidth]{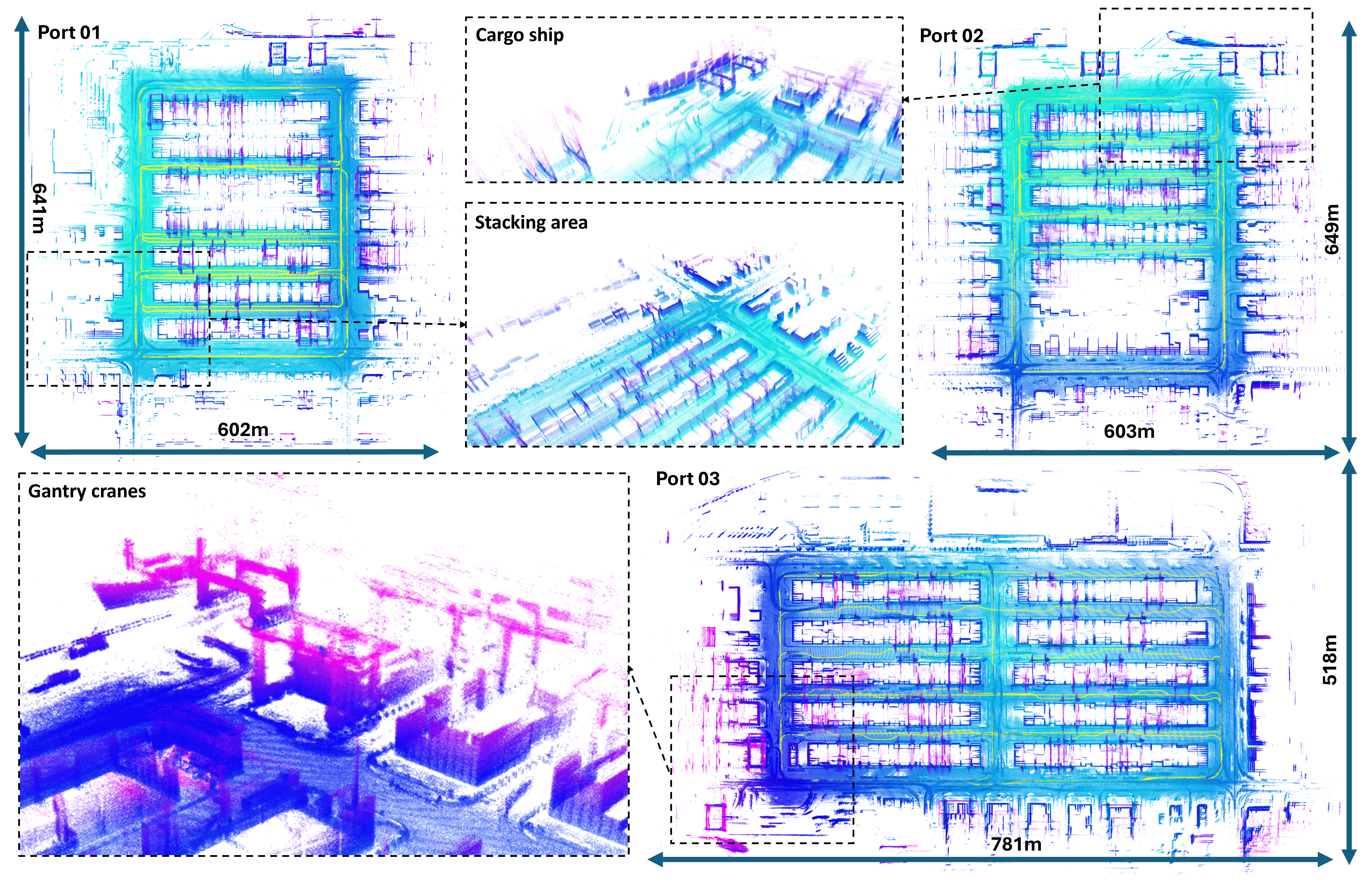}
% \vspace{-1.5em}
\caption{The reconstruction result of CTE-MLO in a port scenario.}\label{Fig: port Mapping}
% \vspace{-2em}
\vspace{-1em}
\end{figure*}
\begin{figure*}[!t]
    \centering	
    \subfigure[Port 01]{
        \includegraphics[width=0.98\linewidth]{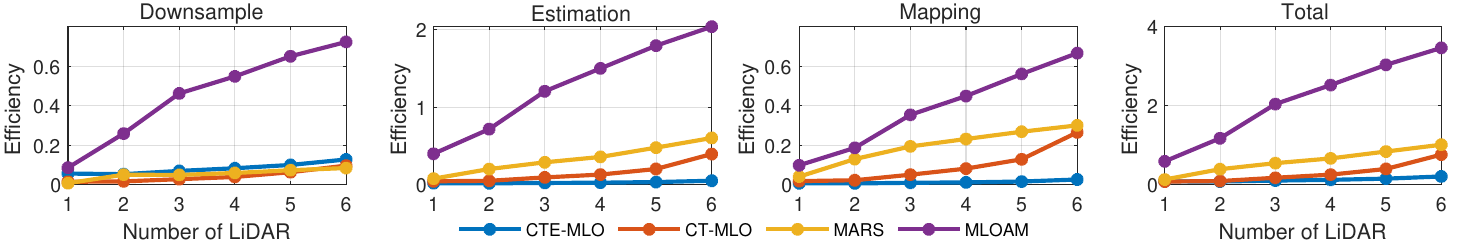}
    }
    % \quad
    \subfigure[Port 02]{
        \includegraphics[width=0.98\linewidth]{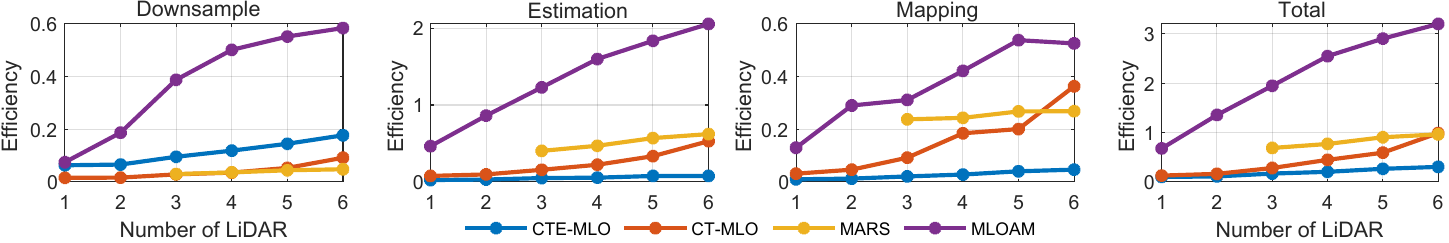} 
        \label{Fig: time_curve_port_02}
    }
    \subfigure[Port 03]{
        \includegraphics[width=0.98\linewidth]{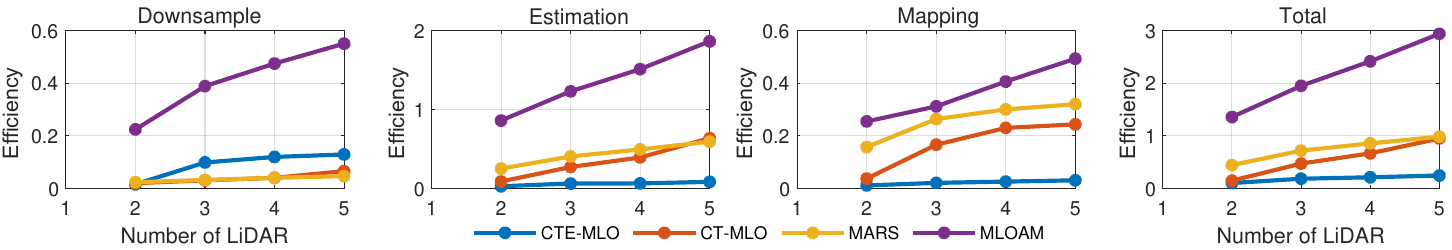} 
        \label{Fig: time_curve_port_03}
    }
    \caption{Benchmark results of efficiency for each individual component with different numbers of LiDARs in large-scale port scenario.} \label{Fig: time_curve_port}
    \vspace{-1.em}
\end{figure*}
% \vspace{-2em}
\subsubsection{Truck Localization in Large-scale Port Scenario} \label{Sec: Truck Localization in Large-scale Port Scenario}
To attest to the practicality, we applied the CTE-MLO to an autonomous driving truck and conducted over $32 km$ of driving tests in three distinct areas within a large-scale port scenario.
The total area of the three test regions reaches $1.182 km^2$.
As shown in Fig. \ref{Fig: Prime Mover}, the truck is equipped with 4 mechanical LiDAR (1 Ouster OS1 and 3 Robosense Helios 32) and 3 Robosense M1 solid-state LiDAR.
Port scenarios are challenging for SLAM methods due to the presence of numerous moving objects (such as trucks and container cranes) and point cloud occlusions.
Thanks to the decentralized multi-LiDAR synchronization strategy proposed in Section \ref{Sec: Decentralized Multi-LiDAR Synchronization}, CTE-MLO can merge LiDAR points sampled from different LiDAR into one scan to provide panorama observation.
The high-quality point cloud shown in Fig. \ref{Fig: port Mapping} illustrates that the proposed method can provides robust trajectory estimation to reconstruct a dense 3D, high-precision map in a large-scale port scenario.
% \balance  

Fig. \ref{Fig: time_curve_port} reports the benchmark results of efficiency for each individual component when performing the large-scale reconstruction with different numbers of LiDARs.
As mentioned in Section \ref{Sec: Running time evaluation}, the efficiency (defined in (\ref{Eq: Efficiency})) is measured by the ratio between processing time and the time interval ($\Delta t=10ms$ for CT-MLO, and CTE-MLO; $\Delta t=50ms$ for MLOAM\cite{MLOAM} and MARS\cite{MARS}).
{In the sequence \textit{Port 02} (Fig. \ref{Fig: time_curve_port_02}), the results of MARS\cite{MARS} are absent when the number of LiDAR is less than 3.
This is attributed to the failure of the estimation due to insufficient constraints.
For the sequence \textit{Port 03} (Fig. \ref{Fig: time_curve_port_03}), all the benchmark methods failed when using a single LiDAR observation and worked fine when the number of LiDARs is greater than or equal to two, which demonstrates the effectiveness of MLO.}
Thanks to the Gaussian process Kalman filter derived in Section \ref{Sec: Continuous-Time LiDAR Odometry}, which enables continuous-time trajectory estimation in only a few linear iterations, CTE-MLO and CT-MLO outperform both MLOAM\cite{MLOAM} and MARS\cite{MARS} in total efficiency.
% Moreover, 
For estimation and mapping, the real-time performance of CT-MLO, MLOAM\cite{MLOAM}, and MARS\cite{MARS} decreases as the number of LiDARs increases due to the large number of point clouds generated by multiple LiDARs worsens the computational complexity.
The localizability-aware point cloud sampling strategy adopted by CTE-MLO reduces the computational complexity of estimation and mapping with the consideration of localizability, which makes the estimation and mapping efficiency of CTE-MLO nearly irrelevant to the number of LiDARs.
As shown in Fig. \ref{Fig: port-CloudMerge}, a large number of redundant points are eliminated by the localizability-aware point cloud sampling strategy.
With the integration of localizability-aware point cloud sampling and continuous-time filter-based registration, CTE-MLO achieves a demonstratively competitive efficiency when compared with SOTA methods.
{The average efficiency of CTE-MLO for processing with 5-6 LiDAR scans in sequences \textit{Port 01}, \textit{Port 02}, and \textit{Port 03} are $0.208 \ll 1.0$, $0.305 \ll 1.0$, and $0.243 \ll 1.0$, demonstrating that the proposed localization and mapping system achieves almost 3-5 times real-time performance without dropping any scan.}
\subsubsection{Autonomous Sweeping in Campus Scenario} \label{Sec: Autonomous Sweeping in Campus Scenario}
\begin{figure*}[!t]
    \centering	
    \subfigure[Campus 01]{
        \includegraphics[width=0.98\linewidth]{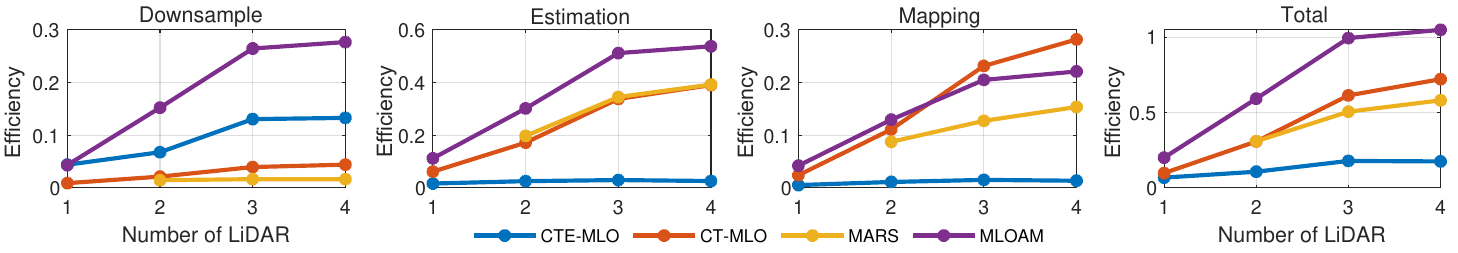}
    }
    \subfigure[Campus 02]{
        \includegraphics[width=0.98\linewidth]{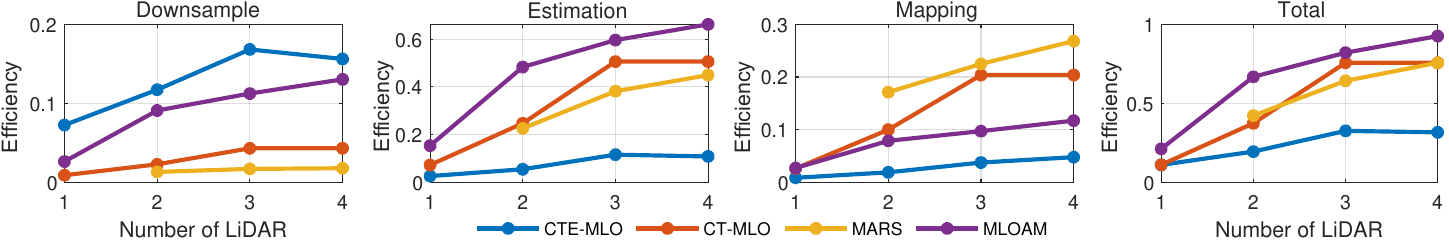} 
    }
    \subfigure[Campus 03]{
        \includegraphics[width=0.98\linewidth]{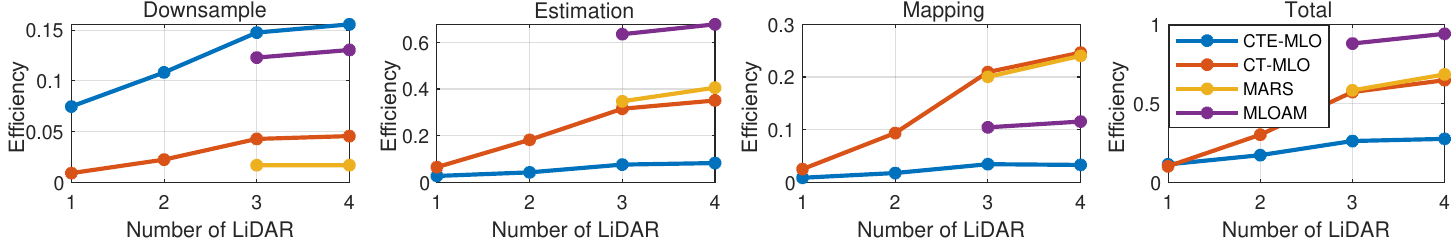} 
    }
    \caption{Benchmark results of efficiency for each individual component with different numbers of LiDARs when performing autonomous sweeping.} \label{Fig: time_curve_sweeper}
\end{figure*}
\begin{figure*}[!t]
    \centering	
    \subfigure[Campus 01]{
        \includegraphics[width=0.3\linewidth]{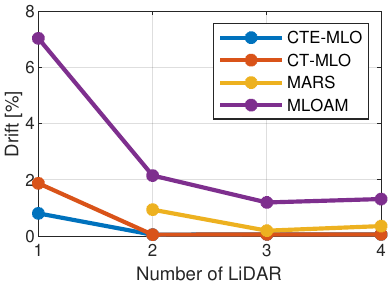}
    }
    % % \quad
    \subfigure[Campus 02]{
        \includegraphics[width=0.3\linewidth]{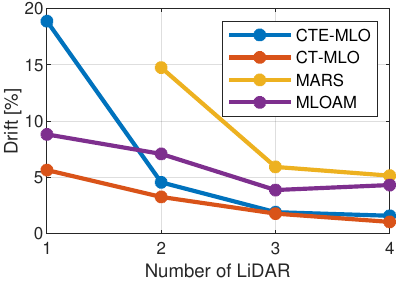} 
    }
    \subfigure[Campus 03]{
        \includegraphics[width=0.3\linewidth]{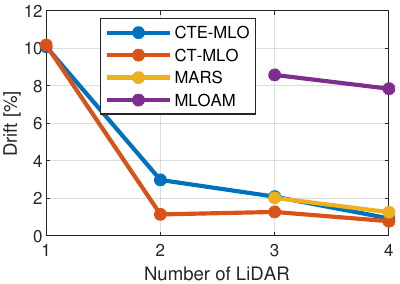} 
    }
    \caption{Benchmark results of accuracy with different numbers of LiDARs when performing autonomous sweeping.} \label{Fig: acc_curve_sweeper}
    % \vspace{-2em}
\end{figure*}
\begin{figure*}[!t]\centering
\includegraphics[width=\linewidth]{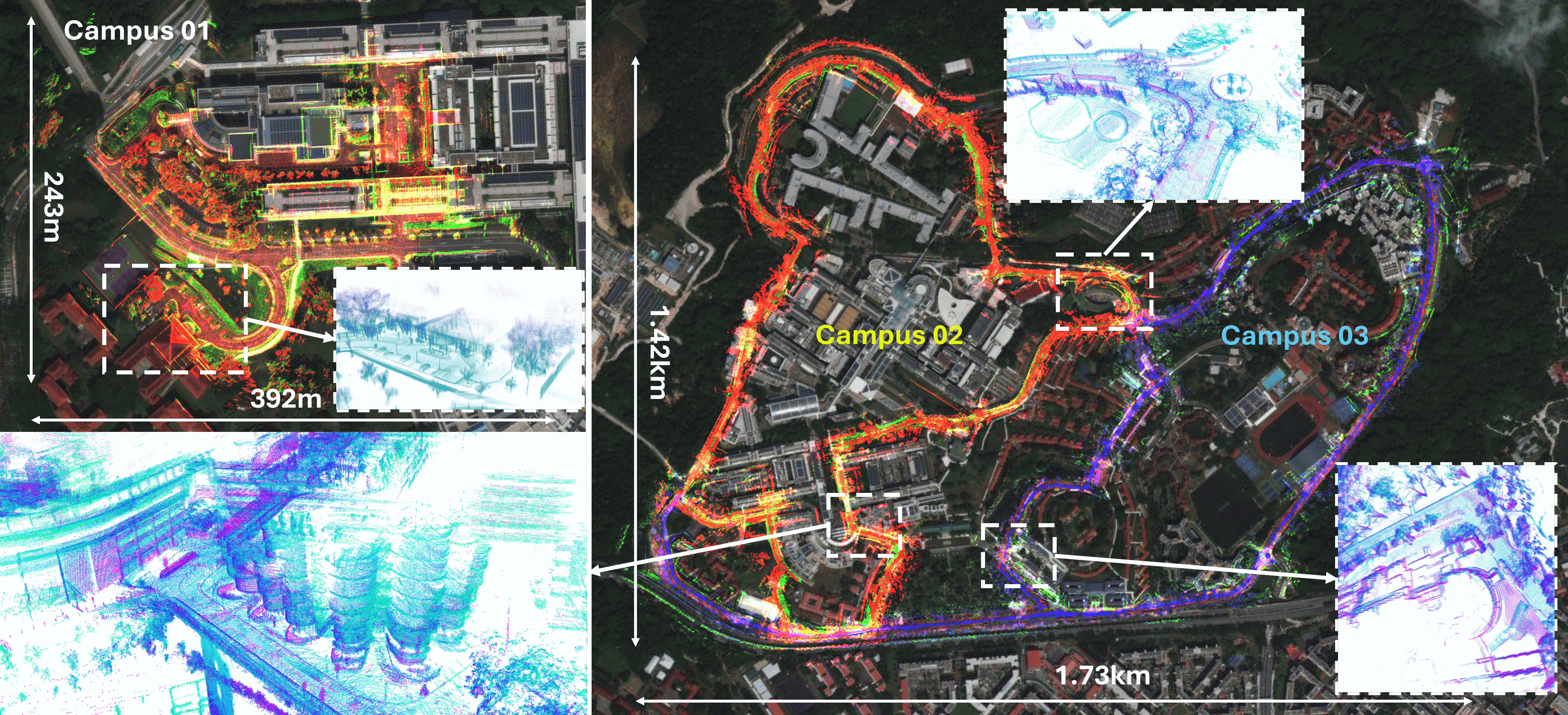}
% \vspace{-1em}
\caption{The reconstruction result of the CTE-MLO during autonomous sweeping.}
\label{Fig: NTU sweeper Mapping}
% \vspace{-1em}
\end{figure*}
As shown in Fig. \ref{Fig: sweeper}, the sweeper is equipped with 3 mechanical LiDARs (1 Robosense Helios 16 and 2 Robosense Helios 32) and 1 Robosense Bpearl solid-state LiDAR.
The relationship between the performance (real-time performance and accuracy) achieved by the proposed method and SOTA methods with respect to the number of LiDAR is demonstrated in Fig. \ref{Fig: time_curve_sweeper} and Fig. \ref{Fig: acc_curve_sweeper}, respectively.
For a fair comparison, the real-time performance is measured by the efficiency defined in (\ref{Eq: Efficiency}), and the accuracy is measured by drift (ratio between relative translation error and total distance).
Due to the numerous buildings on campus, GNSS suffers from the multi-path effect and is inaccurate most of the time, inspired by \cite{tiers,tiers-en}, we collect a prior map of the initial area (container loading area) of the sweeper.
The relative translation error is calculated by registering the LiDAR scan with the prior map when the sweeper returns to the initial area.
For the sequence \textit{Campus 01}, 
the drifts of both CT-MLO and CTE-MLO are well-bounded by $0.06\%$ when the number of LiDARs is greater than or equal to two.
Thanks to the localizability-aware point cloud sampling strategy, only points that have a high contribution to localizability are adopted in CTE-MLO, which enables CTE-MLO to achieve adjacent accuracy to CT-MLO with almost no sacrifice in real-time performance when the number of LiDAR increases.
For sequences \textit{Campus 02} and \textit{Campus 03}, it can be seen that as the number of LiDARs increases, CTE-MLO achieves a significant improvement in accuracy with very little sacrifice in real-time performance.
For sequence \textit{Campus 03}, the results of MLOAM\cite{MLOAM} and MARS\cite{MARS} are absent when the number of LiDAR is less than 3. This is attributed to a large number of LiDAR points being blocked by the sweeper, which leads to estimation degeneration.
CTE-MLO realizes stable localization and mapping under different numbers (1-4) and configurations (mechanical and solid-state) of LiDAR.
{The relative drift of CTE-MLO achieves $0.06\%$ and $1.5\%$ when using 4 LiDARs in mid-scale (\textit{Campus 01}) and large-scale scenarios (\textit{Campus 02, Campus 03}), respectively.
% which outperforms both MLOAM \cite{MLOAM} and MARS\cite{MARS}.
For the real-time performance, the average efficiency of CTE-MLO for processing with 4 LiDAR in sequence \textit{Campus 01}, \textit{Campus 02}, and \textit{Campus 03} are $0.175\ll1.0$, $0.316\ll1.0$, and $0.275\ll1.0$, which is $2$-$3$ times faster than MARS\cite{MARS}, and $3$-$5$ times faster than MLOAM\cite{MLOAM}.}
As shown in Fig. \ref{Fig: NTU sweeper Mapping}, the point cloud reconstructed by CTE-MLO is well-aligned with the satellite image, and the details of buildings and roads are clear.
% \vspace{-1em}
\subsubsection{Autonomous MAV for Field Forest Exploration} \label{Sec: Autonomous MAV for Field Forest Exploration}
\begin{figure*}[!t]\centering
\includegraphics[width=\linewidth]{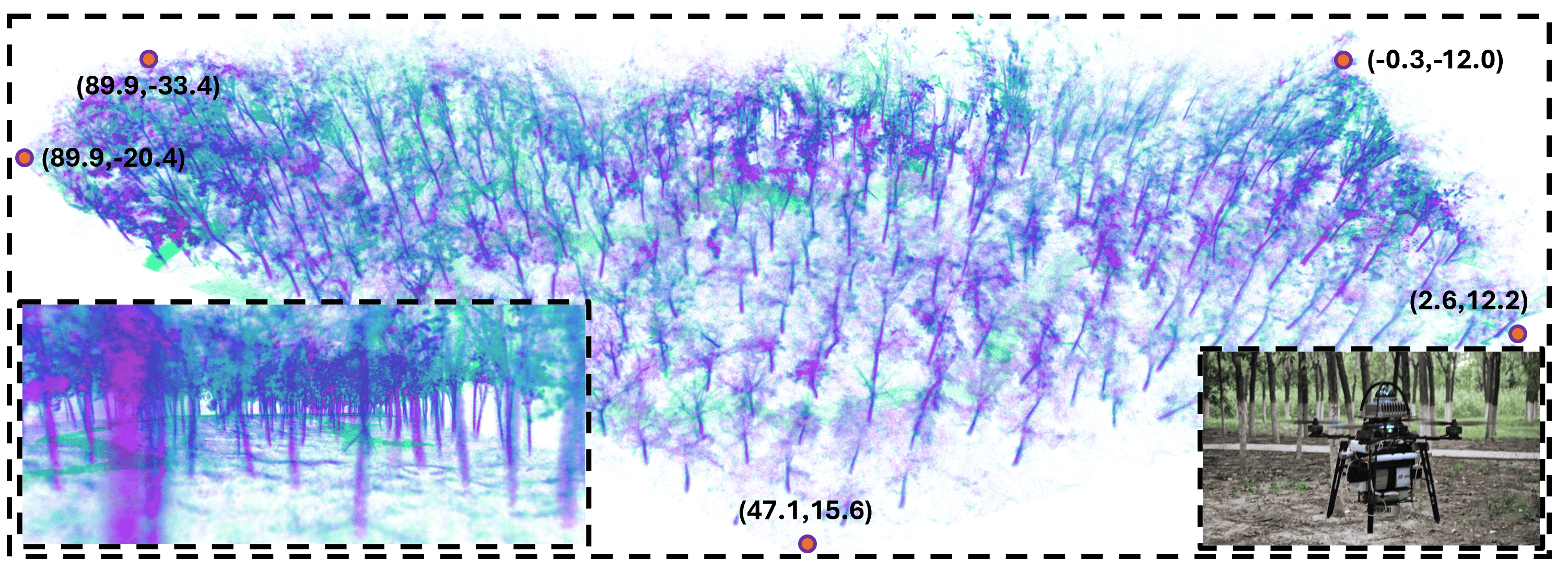}
% \vspace{-1em}
\caption{Point cloud map reconstructed through the autonomous MAV exploration.}
\label{Fig: map_forest}
% \vspace{-1em}
\end{figure*}
\begin{figure}[!t]\centering
\includegraphics[width=\linewidth]{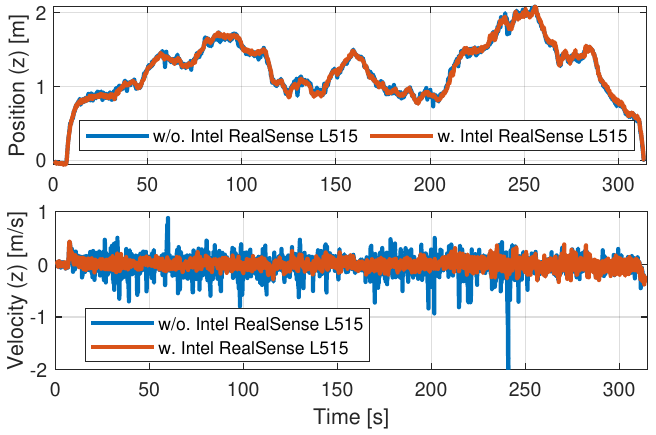}
% \vspace{-2em}
\caption{Comparison results of position and velocity estimation (in z-axis) with or without Intel Realsense L515 LiDAR observation.}
\label{Fig: UAV_pos_vel_plot}
% \vspace{-1em}
\end{figure}

To assess the practicality of the proposed method, we embedded the CTE-MLO into a MAV, enabling an autonomous exploration in a forest. As shown in Fig. \ref{Fig: MAV}, the MAV platform is equipped with two solid-state LiDAR (Livox Mid 360 LiDAR and Intel Realsense L515) and a low-power onboard computer (Nvidia Orin NX). During the autonomous exploration, CTE-MLO is employed to provide real-time trajectory estimation using data from two solid-state LiDARs and coordinate with the planning and control module. The planning and control module is in charge of generating and tracking efficient, collision-free trajectories using the model predictive path integral\cite{williams2017model} and adaptive model predictive control (details can be found in our previous works DPPM\cite{DPPM} and FxTDO-MPC\cite{liwen2024fixedtimedisturbanceobserverbasedmpc}).
As shown in Fig. \ref{Fig: map_forest}, the MAV completed high-precision scanning in a dense forest environment. The area of the exploration regions reaches $2563.66 m^2$.
During the exploration, the MAV platform can block the point cloud observed by Livox Mid 360, leading to a lack of ground observations in the LiDAR odometry. 
To address this issue, we install the RealSense L515 downward to provide point cloud measurements on the ground.
As depicted in Fig. \ref{Fig: UAV_pos_vel_plot}, the proposed CTE-MLO can fully utilize observations from Livox Mid 360 and Intel RealSense L515 point clouds, enabling a more stable estimation of altitude and z-axis velocity compared to single LiDAR odometry.
\section{Conclusion and Future Work}
In this paper, a continuous-time and efficient MLO, called CTE-MLO, is developed to achieve accurate localization and mapping at high frequencies by utilizing observations from multiple LiDARs.
With the detail derivation, we combine the Gaussian process estimation with the Kalman filter which achieves continuous-time trajectory estimation instead of conventional discrete-time state estimation.
The analytic Jacobians for the continuous-time LiDAR update are also derived.
Empowered by the continuous-time trajectory estimation, CTE-MLO enables each LiDAR point in a point stream to query the corresponding continuous-time trajectory using its time stamp and
handles the streaming of continuous-time LiDAR points using non-rigid registration without undistorting points to a certain instant (as noted in Remark \ref{rem: continue time}).
Benefiting from the nature of continuous-time LiDAR measurements, a decentralized multi-LiDAR synchronization scheme is designed to merge and synchronize point clouds from different LiDARs using a LiDAR splitting technique.
The proposed synchronization scheme is totally decentralized without the prior definition of the primary LiDAR and can update the synchronized LiDAR scan with a fixed frequency.
Moreover, considering the trade-off between real-time performance and accuracy in the MLO system, a point cloud sampling technique is developed, which significantly improves the real-time performance of the CTE-MLO with almost no sacrifice of accuracy.
The extensive real-world validations of CTE-MLO, across industrial to field scenarios, demonstrate its effectiveness as a generalized continuous-time MLO. CTE-MLO remains robust under extremely challenging conditions, such as LiDAR degeneration or arbitrary LiDAR data loss, while providing accurate real-time localization and mapping for mobile robots.
The combination of Gaussian process regression and the Kalman filter for continuous-time trajectory estimation enhances the extendability of CTE-MLO, enabling seamless deployment across diverse platforms with heterogeneous LiDAR configurations.
Furthermore, the localizability-aware point cloud sampling ensures nearly LiDAR-number-irrelevant real-time performance for estimation and mapping, enabling CTE-MLO to support autonomy on diverse platforms, from large-size vehicles (e.g., full-size trucks and sweepers) to power-limited agility platforms such as MAVs demonstrated in Section \ref{Sec: Autonomous MAV for Field Forest Exploration}.
With the successful integration of a series of theory and practical implementations (e.g. decentralized synchronization, localizability-aware point cloud sampling, continuous-time registration, and voxel map management) into the Kalman filter, the proposed method is demonstratively competitive compared to SOTA MLO methods.

Currently, thanks to the tightly coupled fusion of multi-LiDAR observations, CTE-MLO achieves accurate and robust trajectory estimation in real-world scenarios. However, extreme weather conditions (such as heavy rain, and fog) may impact the robustness of LiDAR odometry. Leveraging the exceptional extendability of continuous-time estimation (as noted in Remark \ref{rem: Extendability}), future work could consider integrating CTE-MLO with additional sensors (e.g., radar, thermal cameras, GNSS, wheel odometry) that are not sensitive to the extreme weather.
The proposed continuous-time estimation allows observations from different sensors to constrain the continuous-time trajectory, making it easy to incorporate additional sensors into CTE-MLO without requiring strict synchronization.
The extensibility of CTE-MLO presents significant potential for developing a robust and full-fledged tightly coupled multi-sensor fusion localization and mapping system capable of adapting to diverse weather conditions.

% \section*{ACKNOWLEDGMENT}
% This research is supported by the National Research Foundation (NRF), Singapore, under the NRF Medium Sized Centre scheme.

\bibliographystyle{Bibliography/IEEEtran}
\bibliography{Bibliography/tj-template-ap}\ %IEEEabrv instead of IEEEfull
\vspace{-1em}
\begin{IEEEbiography}[{\includegraphics[width=1in,height=1.25in,clip,keepaspectratio]{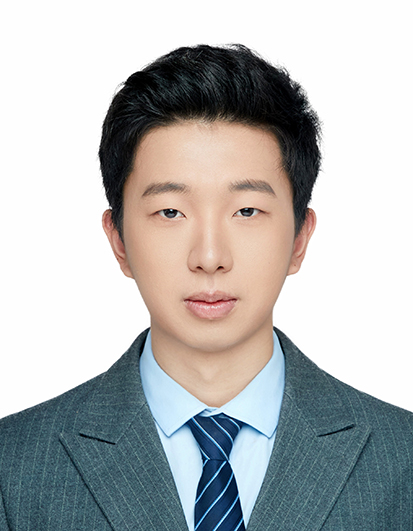}}]
{Hongming Shen}~is currently a Research Fellow with the Centre for Advanced Robotics Technology Innovation (CARTIN), Nanyang Technological University, Singapore. Before joining NTU, he received his Ph.D. degree in Control Theory and Control Engineering from Tianjin University, Tianjin, China, in 2023. He has published several papers in prestigious journals/conferences, including IEEE T-IE, IEEE/ASME T-MECH, IEEE T-IM, IEEE T-IV, RA-L, CVPR, ICRA, and IROS. He has also served as a reviewer in several well-acknowledged journals, such as IEEE T-FR, IEEE T-IE, Satellite Navigation, and RA-L. He received IEEE ICARCV Best Paper Award in 2024, and 1st Prize in the Aerial Swarm Challenge in 2022. His current research interests include state estimation, multi-sensor fusion, localization and mapping, and aerial Robotics.
\end{IEEEbiography}
\vspace{-1em}
\begin{IEEEbiography}[{\includegraphics[width=1in,height=1.25in,clip,keepaspectratio]{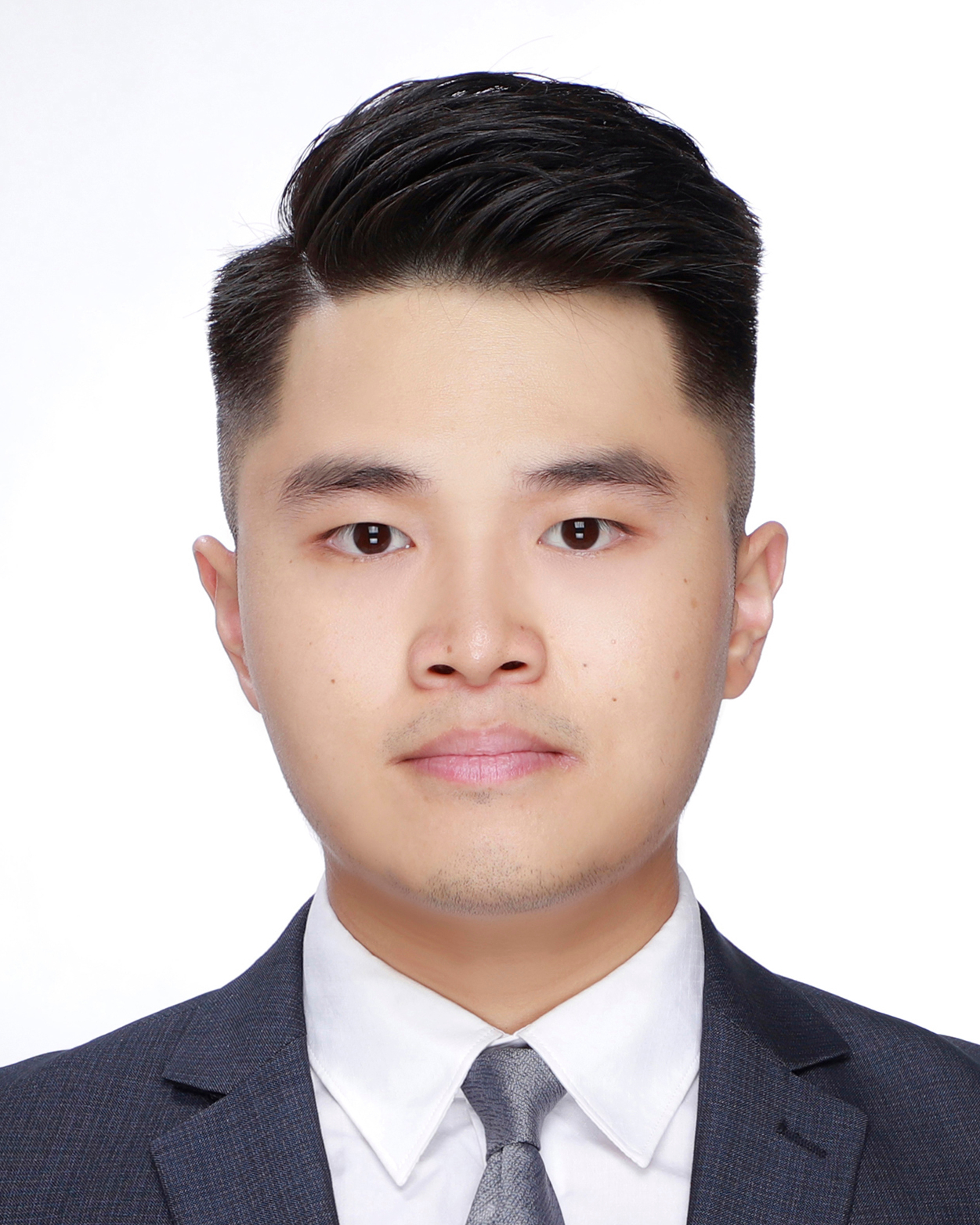}}]
{Zhenyu Wu}~received his B.Eng. degree of Electrical Engineering and Automation from Wuhan University, China, in 2016, the M.Sc. and Ph.D. degrees from Nanyang Technological University (NTU), Singapore, in 2017 and 2022, respectively.

He is currently a Research Assistant Professor with the Centre for Advanced Robotics Technology Innovation (CARTIN), NTU. He has published several papers in prestigious journals/conferences, including IEEE/ASME T-MECH, IEEE T-IM, ICRA, IROS, CVPR, and ITSC. In addition, he has also served as a reviewer in a number of well-acknowledged journals and conferences, such as IEEE T-IE, T-GRS, T-IM, RA-L, ICRA, and IROS. He received 2024 IEEE ICARCV Best Paper Award. His research interests include intelligent perception, localization, and navigation for autonomous systems in complex environments.
\end{IEEEbiography}
\vspace{-1em}
\begin{IEEEbiography}
[{\includegraphics[width=1in,height=1.25in,clip,keepaspectratio]{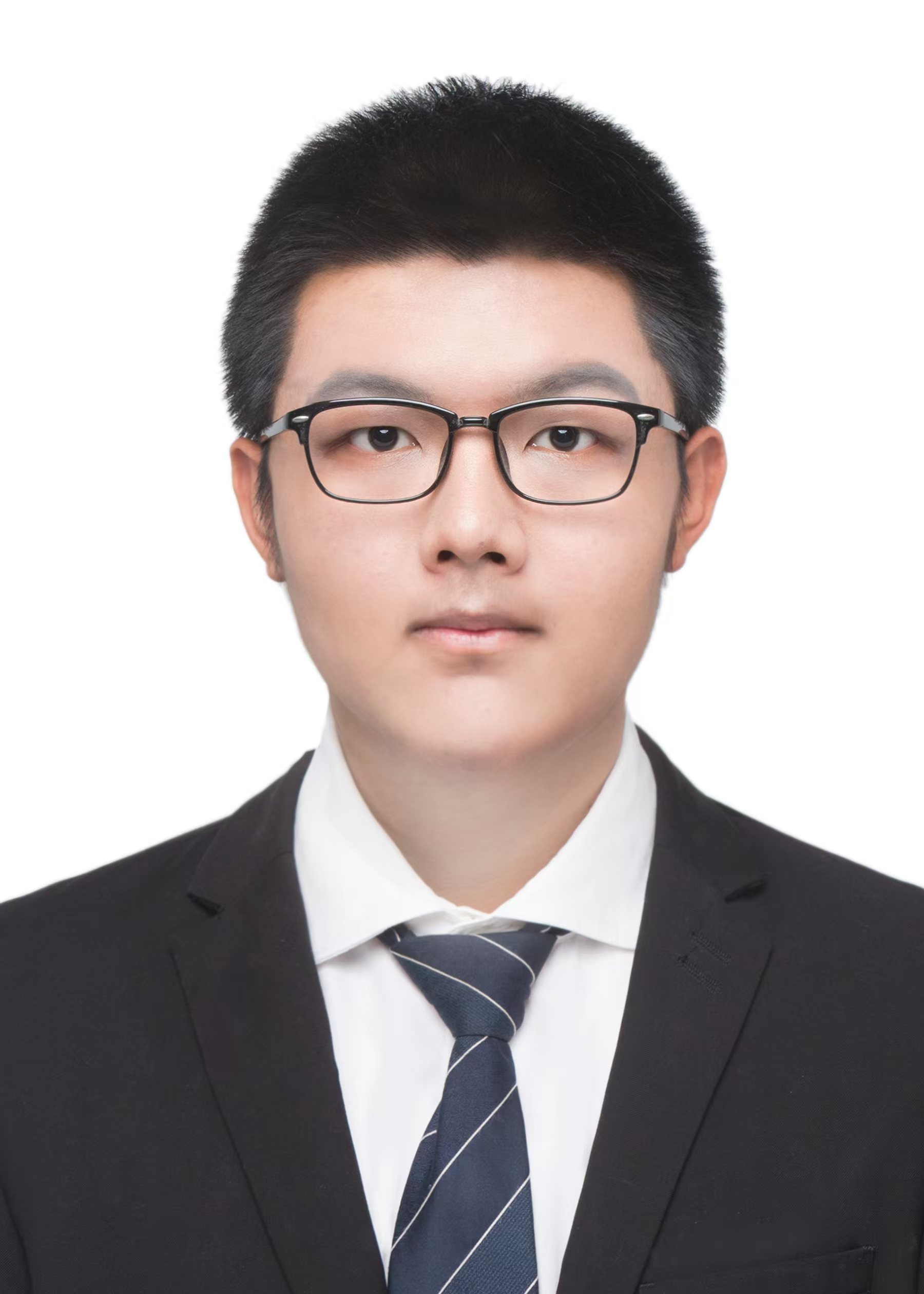}}]
% {Yulin Hui}~received the B.S. degree from the School of Electrical and Information Engineering, Tianjin University, Tianjin, China, in 2021. He is currently working toward the Ph.D. degree in control theory and control engineering with Tianjin University. His research interests include aerial robotics, autonomous exploration, and trajectory planning.
{Yulin Hui}~received a B.S. degree in Automation from the School of Electrical and Information Engineering, Tianjin University, Tianjin, China, in 2021. He is currently working toward a Ph.D. degree in control theory and control engineering at Tianjin University, China. He received 1st Prize in the Aerial Swarm Challenge in 2022. His main research interests include autonomous aerial robotics, multi-UAV cooperative exploration in complex environments, and fast trajectory planning in clustered environments.
\end{IEEEbiography}
% \vspace{-1em}
\begin{IEEEbiography}
[{\includegraphics[width=1in,height=1.25in,clip,keepaspectratio]{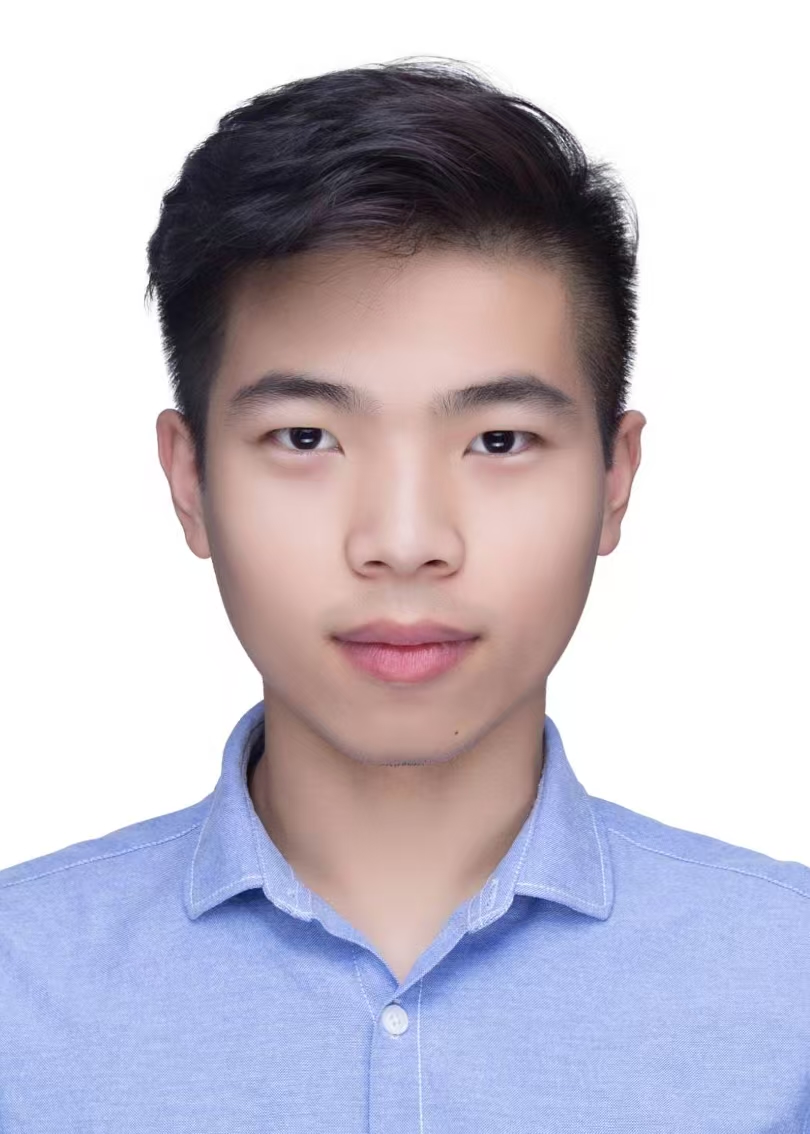}}]
{Wei Wang}~received his M.Sc. degree from the School of Electrical and Electronic Engineering, Nanyang Technological University (NTU), Singapore, in 2022, and his B.Eng. degree in Automation from Zhejiang University, Hangzhou, China, in 2020. He is currently pursuing a doctoral degree at the School of Electrical and Electronic Engineering, NTU, Singapore.
His research interests include multi-sensor fusion for autonomous vehicle localization and navigation in complex terrains and GPS-denied environments, as well as intelligent transportation systems.
\end{IEEEbiography}
% \vspace{-1em}
\begin{IEEEbiography}
[{\includegraphics[width=1in,height=1.25in,clip,keepaspectratio]{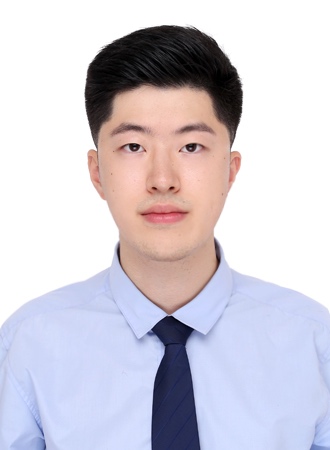}}]
{Qiyang Lyu}~received his B.Eng. degree in Electronic Information Engineering from the University of Electronic Science and Technology of China, Chengdu, China, in 2020, and his M.Sc. degree in Computer Control and Automation from the School of Electrical and Electronic Engineering, Nanyang Technological University (NTU), Singapore, in 2021. He is currently working toward a Ph.D. degree within the School of Electrical and Electronic Engineering, NTU, Singapore. His research interests include sensor calibration, multi-modal mapping and localization for unmanned vehicles. 
\end{IEEEbiography}
\vspace{-1.5em}
\begin{IEEEbiography}
[{\includegraphics[width=1in,height=1.25in,clip,keepaspectratio]{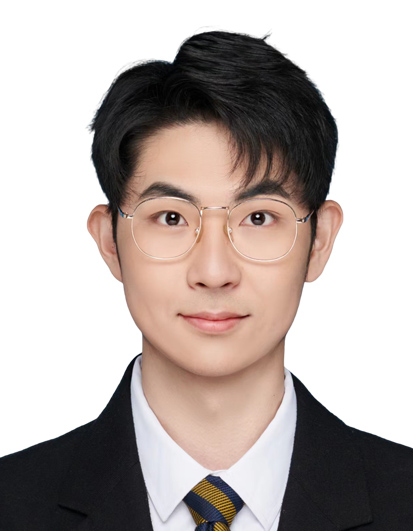}}]
{Tianchen Deng}~received the B.Eng. degree in control science and engineering from the Harbin Institute of Technology,
Harbin, China, in 2021. He is currently pursuing the Joint
Ph.D. degree in control science and engineering with
Shanghai Jiao Tong University, Shanghai, China and Nanyang Technological University, Singapore.
His main research interests include visual simultaneous localization and mapping (VSLAM), vision-based localization, 3D Reconstruction and navigation of autonomous robot.
\end{IEEEbiography}
\vspace{-1.5em}
\begin{IEEEbiography}
[{\includegraphics[width=1in,height=1.25in,clip,keepaspectratio]{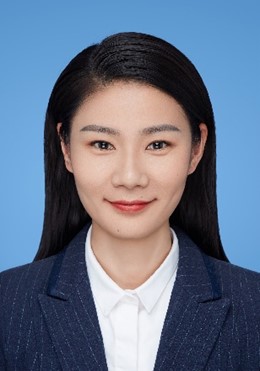}}]
{Yeqing Zhu}~is currently a Postdoctoral Research Fellow with the School of Aerospace Engineering, Beijing Institute of Technology. She is also a visiting scholar with the Centre for Advanced Robotics Technology Innovation (CARTIN), Nanyang Technological University (NTU), Singapore. She received a Ph.D. degree from Beijing Institute of Technology, Beijing, China, in 2022. Her main research interests include computer vision, simultaneous localization and mapping for autonomous robots, flight dynamics, and control.
\end{IEEEbiography}
\vspace{-1.5em}
\begin{IEEEbiography}
[{\includegraphics[width=1in,height=1.25in,clip,keepaspectratio]{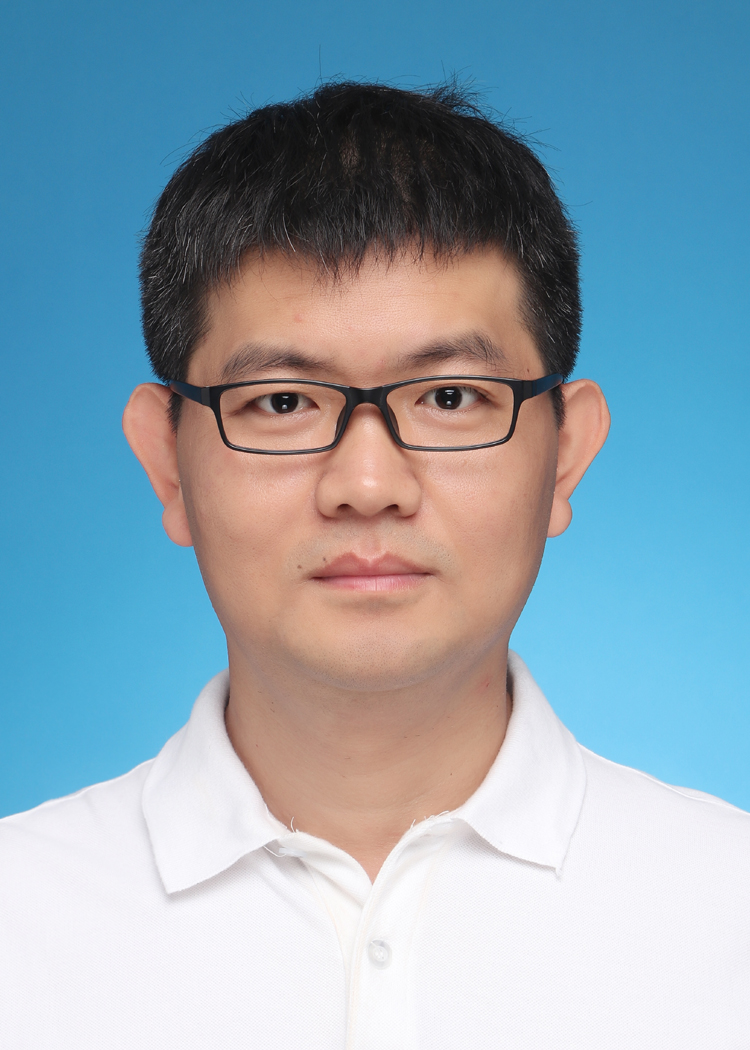}}]
{Bailing Tian}~received the B.S., M.S. and Ph.D degrees in automatic control from Tianjin University, Tianjin, China, in 2006, 2008, and 2011, respectively.
He was an academic visitor at the School of Electrical and Electronic Engineering, University of Manchester from June 2014 to June 2015. He is currently a professor at the School of Electrical and Information Engineering, Tianjin University. His main research interests include finite time control, motion planning, simultaneous localization and mapping, and integrated guidance and control for unmanned systems.
\end{IEEEbiography}
\vspace{-1.5em}
\begin{IEEEbiography}
[{\includegraphics[width=1in,height=1.25in,clip,keepaspectratio]{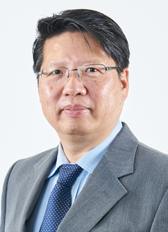}}]{Danwei Wang}~(Life Fellow, IEEE) received his Ph.D. and M.S.E. degrees from the University of Michigan, Ann Arbor, USA, in 1989 and 1984, respectively. He received his B.E degree from the South China University of Technology, China, in 1982. 
	
Since 1989, he has been with the School of Electrical and Electronic Engineering, Nanyang Technological University, Singapore. Currently, he is Full Professor and the Programme Director of the Centre for Advanced Robotics Technology Innovation (CARTIN). He is the Chair of IEEE Singapore Robotics and Automation Chapter and a senator in NTU Academics Council. He has served as general chairman, technical chairman and various positions in international robotics and control conferences, such as ICRA, IROS, and ICARCV conferences. He is an associate editor for the International Journal of Humanoid Robotics and served as an associate editor of Conference Editorial Board, IEEE Control Systems Society from 1998 to 2005. He has published 5 books and over 500 technical articles in the areas of robotics and control theories, as well as fault diagnosis and prognosis of complex engineering systems. He was a recipient of Alexander von Humboldt Fellowship, Germany, and ST Engineering Distinguished Professor Award, Singapore. He is a Fellow of the Academy of Engineering, Singapore (SAEng) and Life Fellow of the IEEE. His research interests include robotics, control theory and applications. 
\end{IEEEbiography}
% \balance
\vfill\pagebreak
\end{document}